%% file: paper_iclr.tex
\documentclass{article} 
\usepackage{iclr2018_conference,times}
\usepackage{hyperref}
\usepackage{url}
\usepackage{booktabs}

\usepackage{soul} 
\input{notation.tex}
\usepackage{nicefrac}

\title{Learning a generative model for validity in complex discrete structures}

\author{David Janz \\
University of Cambridge\\
\texttt{dj343@cam.ac.uk}
\And
Jos van der Westhuizen \\
University of Cambridge\\
\texttt{jv365@cam.ac.uk}
\And
Brooks Paige\\
Alan Turing Institute\\
University of Cambridge \\
\texttt{bpaige@turing.ac.uk}
\And
Matt J. Kusner\\
Alan Turing Institute\\
University of Warwick \\
\texttt{mkusner@turing.ac.uk}
\And
Jos{\'e} Miguel Hern{\'a}ndez-Lobato\\
Alan Turing Institute\\
University of Cambridge\\
\texttt{jmh233@cam.ac.uk}
}

\iclrfinalcopy 

\begin{document}

\maketitle

\begin{abstract}
  Deep generative models have been successfully used to learn representations
  for high-dimensional discrete spaces by representing discrete objects as
  sequences and employing powerful sequence-based deep models. Unfortunately,
  these sequence-based models often produce invalid sequences: sequences which
  do not represent any underlying discrete structure; invalid sequences hinder
  the utility of such models. As a step towards solving this problem, we propose
  to learn a deep recurrent validator model, which can estimate whether a
  partial sequence can function as the beginning of a full, valid sequence. This
  validator provides insight as to how individual sequence elements influence
  the validity of the overall sequence, and can be used to constrain sequence
  based models to generate valid sequences -- and thus faithfully model discrete
  objects. Our approach is inspired by reinforcement learning, where an oracle
  which can evaluate validity of complete sequences provides a sparse reward
  signal. We demonstrate its effectiveness as a generative model of Python 3
  source code for mathematical expressions, and in improving the ability of a
  variational autoencoder trained on SMILES strings to decode valid molecular
  structures.
\end{abstract}

\section{Introduction}
\input{sections/introduction}

\section{A model for sequence validity}
\input{sections/model_description}

\section{Online generation of synthetic training data}
\input{sections/active_learning}

\section{Experiments}
\input{sections/python}
\input{sections/molecules}

\section{Discussion}
\input{sections/discussion}


\bibliography{references_iclr}
\bibliographystyle{iclr2018_conference}

\newpage
\appendix
\label{appendix}
\input{sections/appendix}
\end{document}

%% file: notation.tex
\usepackage{subcaption}

\usepackage{booktabs} 
\usepackage{multirow}

\usepackage{algorithm}
\usepackage{algorithmic}
\usepackage{amsmath}
\usepackage{amssymb}
\usepackage{mathtools}
\usepackage{pgf}
\usepackage{import}

\usepackage{microtype}      
\usepackage{xfrac}

\newcommand{\inspace}{\mathcal{X}}

\newcommand{\weights}{w}

\newcommand{\set}[1]{\{#1\}}

\DeclareMathOperator*{\argmax}{argmax}
\DeclarePairedDelimiterX{\inner}[2]{\langle}{\rangle}{#1, #2}

\newcommand{\out}{\mathrm{y}(x_t|x_{<t},\weights)}

\newcommand{\valid}{{\tilde v}}

\newcommand{\data}{\mathcal{D}}


%% file: sections/introduction.tex

Deep generative modeling has seen many successful recent developments, such as
producing realistic images from noise \citep{radford_unsupervised_2015} and
creating artwork \citep{gatys2016image}. We find particularly promising the
opportunity to leverage deep generative models for search in high-dimensional
discrete spaces \citep{gomez-bombarelli_automatic_2016,kusner17}. Discrete
search is at the heart of problems in drug discovery \citep{gomez2016design},
natural language processing \citep{Bowman2016,Guimaraes2017}, and symbolic
regression \citep{kusner17}.

The application of deep modeling to search involves `lifting' the search from
the discrete space to a continuous space, via an {autoencoder}
\citep{rumelhart1985learning}. An autoencoder learns two mappings: 1) a mapping
from discrete space to continuous space called an \emph{encoder}; and 2) a
reverse mapping from continuous space back to discrete space called a
\emph{decoder}. The discrete space is presented to the autoencoder as a sequence
in some formal language --- for example, in
\citet{gomez-bombarelli_automatic_2016} molecules are encoded as SMILES strings
--- and powerful sequential models (e.g., LSTMs \citep{hochreiter1997long} GRUs
\citep{cho2014learning}, DCNNs \citep{kalchbrenner2014convolutional}) are
applied to the string representation. When employing these models as encoders
and decoders, generation of invalid sequences is however possible, and using
current techniques this happens frequently. \citet{kusner17} aimed to fix this
by basing the sequential models on parse tree representations of the discrete
structures, where externally specified grammatical rules assist the model in the
decoding process. This work boosted the ability of the model to produce valid
sequences during decoding, but its performance achieved by this method leaves
scope for improvement, and the method requires hand-crafted grammatical rules
for each application domain.

In this paper, we propose a generative approach to modeling validity that can
learn the validity constraints of a given discrete space. We show how concepts
from reinforcement learning may be used to define a suitable generative model
and how this model can be approximated using sequence-based deep learning
techniques. To assist in training this generative model we propose two data
augmentation techniques. Where no labeled data set of valid and invalid
sequences is available, we propose a novel approach to active
learning for sequential tasks inspired by classic mutual-information-based
approaches \citep{houlsby2011bayesian,hernandez2014predictive}. In the context
of molecules, where data sets containing valid molecule examples do exist, we
propose an effective data augmentation process based on applying minimal
perturbations to known-valid sequences. These two techniques allow us to rapidly
learn sequence validity models that can be used as a) generative models, which
we demonstrate in the context of Python 3 mathematical expressions and b) a
grammar model for character-based sequences, that can drastically improve the
ability of deep models to decode valid discrete structures from continuous
representations. We demonstrate the latter in the context of molecules
represented as SMILES strings.


%% file: sections/model_description.tex

To formalise the problem we denote the set of discrete sequences of length $T$
by $\inspace = \set{(x_1, \dots, x_T) : x_t \in \mathcal{C}}$ using an alphabet
$\mathcal{C} = \set{1, \dots, C}$ of size $C$. Individual
sequences in $\inspace$ are denoted $x_{1:T}$. We assume the availability of a
\textit{validator} $v \colon \inspace \to \set{0, 1}$, an oracle which can tell
us whether a given sequence is valid. It is important to note that such a
validator gives very sparse feedback: it can only be evaluated on a {\em
  complete} sequence. Examples of such validators are compilers for programming
languages (which can identify syntax and type errors) and chemo-informatics
software for parsing SMILES strings (which identify violations of valence
constraints). Running the standard validity checker $v(x_{1:T})$ on a partial
sequence or subsequence (e.g.,\ the first $t < T$ characters of a computer
program) does not in general provide any indication as to whether the
complete sequence of length $T$ is valid.

\begin{figure}[t]
	\centering
  \begin{minipage}[c]{0.5\textwidth}
	\includegraphics[width=\textwidth]{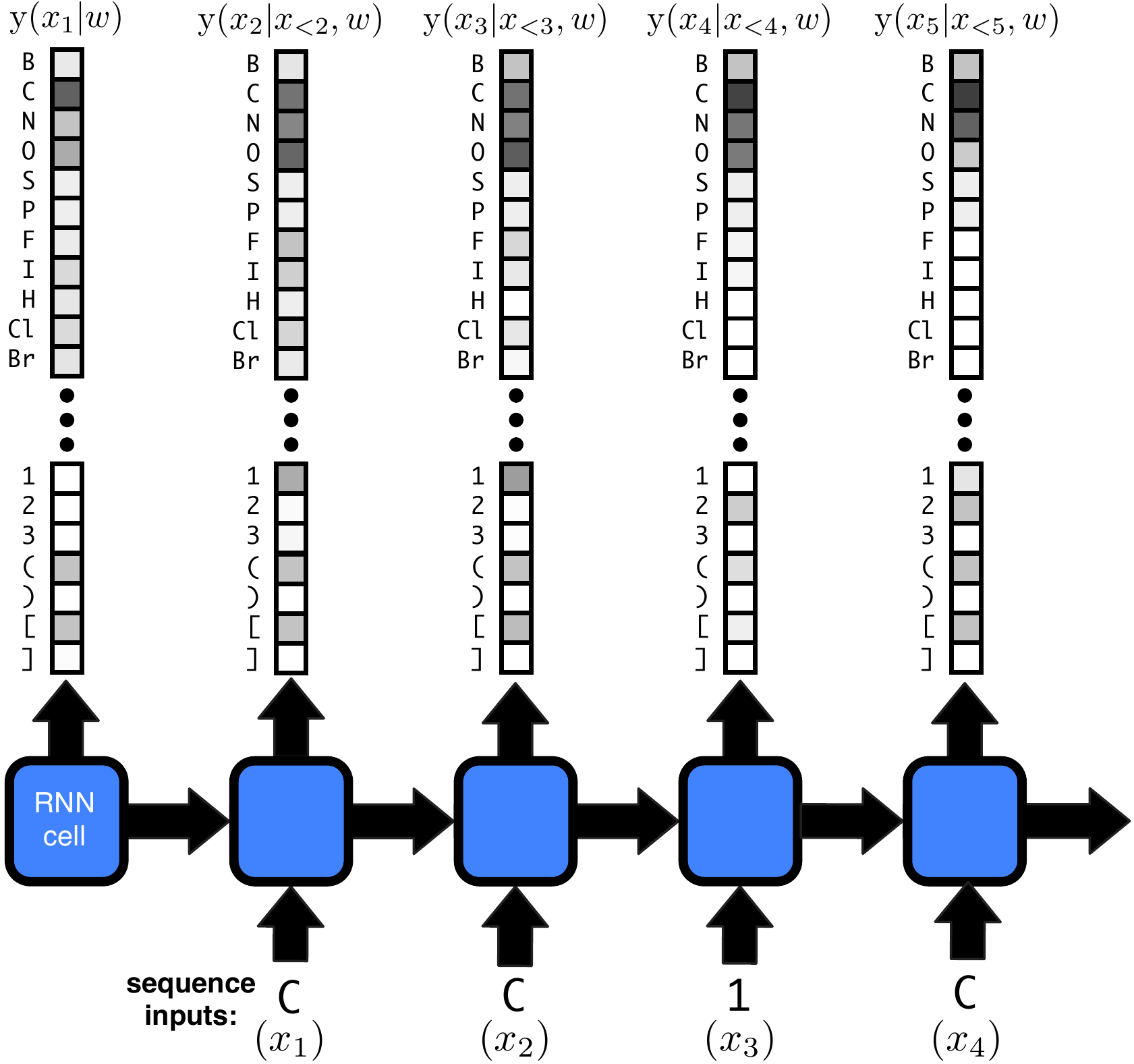}
  \end{minipage}\hfill
\begin{minipage}[c]{0.45\textwidth}\vspace{2em}
  \caption{The recurrent model used to approximate the $ Q $-function. A hypothetical
    logistic function activation is shown for each character in $ \mathcal{C} $.
    Here the set of characters is the SMILES alphabet and we use the first 3
    characters of the molecule in figure \ref{fig:example-mol} as the input
    example. The initial character is predicted from the first hidden state, and
    the LSTM continues until the end of the sequence.}
	\label{fig:rnn_model}
  \end{minipage}\hfill
\end{figure}

We aim to obtain a generative model for the sequence set
$\mathcal{X}_+ = \set{x_{1:T} \in \inspace : v(x_{1:T})=1}$, the subset
of valid sequences in $\inspace$. To achieve this, we would
ideally like to be able to query a more-informative function $\valid(x_{1:t})$
which operates on prefixes $x_{1:t}$ of a hypothetical longer sequence
$x_{1:T}$ and outputs
\begin{align}
\valid(x_{1:t}) &=
\begin{cases}
  1 &\text{if there exists a suffix}\ x_{t+1:T}\ \text{such that}\ v([x_{1:t}, x_{t+1:T}]) = 1,  \\
  0 &\text{otherwise}
\end{cases}
\label{eq:vhat}
\end{align}
where $[x_{1:t}, x_{t+1:T}]$ concatenates a prefix and a suffix to form
a complete sequence. The function $\valid(x_{1:t})$ can be used to determine
whether a given prefix can ever successfully yield a valid outcome. Note that we
are indifferent to {\em how many} suffixes yield valid sequences. With
access to $\valid(x_{1:t})$, we could create a generative model for
$\mathcal{X}_+$ which constructs sequences from left to right, a single
character at a time, using $\valid(x_{1:t})$ to provide early feedback as to
which of the next character choices will surely not lead to a ``dead end'' from
which no valid sequence can be produced.

We frame the problem of modeling $\mathcal{X}_+$ as a Markov decision process
\citep{sutton1998reinforcement} for which we train a reinforcement learning
agent to select characters sequentially in a manner that avoids producing
invalid sequences. At time $t=1,\dots,T$, the agent is in state
$x_{<t} = x_{1:t-1}$ and can take actions $x_t \in \mathcal{C}$. At the end of
an episode, following action $x_T$, the agent receives a reward of $v(x_{1:T})$.
Since in practice we are only able to evaluate $v(x_{1:T})$ in a meaningful way
on complete sequences, the agent does not receive any reward at any of the
intermediate steps $t<T$. The optimal $Q$-function $Q^\star(s, a)$
\citep{watkins1989learning}, a function of a state $s$ and an action $a$,
represents the expected reward of an agent following an optimal policy which
takes action $a$ at state $s$. This optimal $Q$-function assigns value $1$ to
actions $a=x_t$ in state $s=x_{<t}$ for which there exists a suffix $x_{t+1:T}$
such that $[x_{1:t}, x_{t+1:T}] \in \mathcal{X}_+$, and value $0$ to all
other state/action pairs. This behaviour exactly matches the desired prefix
validator in \eqref{eq:vhat}, that is, $Q^\star(x_{<t}, x_t) = \valid(x_{1:t})$,
and so for the reinforcement learning environment as specified, learning
$\valid(x_{1:t})$ corresponds to learning the Q-function.

Having access to $Q^\star$ would allow us to obtain a generative model for
$\mathcal{X}_+$. In particular, an agent following any optimal policy
$\pi^\star(x_{<t})=\argmax_{x_t\in\mathcal{C}}Q^\star(x_{<t}, x_t)$ will always
generate valid sequences. If we sample uniformly at random across all optimal
actions at each time $t=1,\ldots,T$, we obtain the joint distribution given by
\begin{align}
  p(x_{1:T}) = \prod_{t=1}^T \frac{Q^\star(x_{<t}, x_t)}{Z(x_{<t})}, \label{eq:distribution}
\end{align}
where $Z(x_{<t})=\sum_{x_t} Q^\star(x_{<t}, x_t)$ are the per-timestep
normalisation constants. This distribution allows us to sample sequences
$x_{1:T}$ in a straightforward manner by sequentially selecting characters
$x_t \in \mathcal{C}$ given the previously selected ones in $x_{<t}$.

\newcommand{\outhat}{\mathrm{y}(x_t|x_{<t},\hat{\weights})}

In this work we focus on learning an approximation to (\ref{eq:distribution}).
For this, we use recurrent neural networks, which have recently shown remarkable
empirical success in the modeling of sequential data, e.g., in natural language
processing applications \citep{sutskever2014sequence}. We approximate the
optimal $Q$-function with a long-short term memory (LSTM) model
\citep{hochreiter1997long} that has one output unit per character in
$\mathcal{C}$, with each output unit using a logistic activation function (see
figure \ref{fig:rnn_model}), such that the output is in the closed interval
$[0,1]$. We denote by $\out$ the value at time $t$ of the LSTM output unit
corresponding to character $x_t$ when the network weights are $w$ the input is
the sequence $x_{<t}$. We interpret the neural network output $\out$ as
$p(Q^\star(x_{<t}, x_t) = 1)$, that is, as the probability that action $x_t$ can
yield a valid sequence given that the current state is $x_{<t}$.

Within our framing a sequence $x_{1:T}$ will be valid according to our model if every action during
the sequence generation process is permissible, that is, if
$Q^\star(x_{<t}, x_t) = 1$ for $t=1,\ldots,T$. Similarly, we consider that the
sequence $x_{1:T}$ will be invalid if at least one action during the sequence
generation process is not valid\footnote{ Note that, once $Q^\star(x_{<t}, x_t)$
  is zero, all the following values of $Q^\star(x_{<t}, x_t)$ in that sequence
  will be irrelevant to us. Therefore, we can safely assume that a sequence is
  invalid if $Q^\star(x_{<t}, x_t)$ is zero at least once in the sequence.},
that is, if $Q^\star(x_{<t}, x_t) = 0$ at least once for $t=1,\ldots,T$. This
specifies the following log-likelihood function given a training set
$\mathcal{D}=\{ (x^n_{1:T}, y_n) \}_{n=1}^N$ of sequences $x^n_{1:T}\in\inspace$
and corresponding labels $y_n = v(x_{1:T})$:
\begin{align}
\label{eq:loss}
  \mathcal{L}(w|\mathcal{D}) = \sum_{n=1}^N \left\{ y_n \log p(y_n = 1|x_{1:T}^n,w) + (1 - y_n) \log p(y_n = 0|x_{1:T}^n,w) \right\}\,,
\end{align}
where, following from the above characterisation of valid and invalid sequences,
we define
\begin{align}
p(y_n=1|x_{1:T}^n, w) & = \prod_{t=1}^T \out\,, &
p(y_n=0|x_{1:T}^n, w) & = 1- \prod_{t=1}^T \out\,,\label{eq:predictions}
\end{align}
according to our model's predictions. The log-likelihood (\ref{eq:loss}) can be
optimised using backpropagation and stochastic gradient descent and, given
sufficient model capacity, results in a maximiser $\hat{w}$ such that
$\outhat \approx Q^\star(x_{<t},x_t)$.

Instead of directly maximising (\ref{eq:loss}), we can follow a Bayesian
approach to obtain estimates of uncertainty in the predictions of our LSTM
model. For this, we can introduce dropout layers which stochastically zero-out
units in the input and hidden layers of the LSTM model according to a Bernoulli
distribution \citep{gal2016theoretically}. Under the assumption of a Gaussian
prior $p(w)$ over weights, the resulting stochastic process yields an implicit
approximation $q(w)$ to the posterior distribution
$p(w | \data) \propto \exp(\mathcal{L}(w|\mathcal{D}))p(w)$. We do this to
obtain uncertainty estimates, allowing us to perform efficient active learning,
as described in section \ref{sec:active}.


%% file: sections/active_learning.tex

One critical aspect of learning $w$ as described above is how to generate the
training set $\mathcal{D}$ in a sensible manner. A na{\"i}ve approach could be
to draw elements from $\inspace$ uniformly at random. However, in many cases,
$\inspace$ contains only a tiny fraction of valid sequences and the uniform
sampling approach produces extremely unbalanced sets which contain very little
information about the structure of valid sequences. While rejection sampling can
be used to increase the number of positive samples, the resulting additional
cost makes such an alternative infeasible in most practical cases. The problem
gets worse as the length of the sequences considered $T$ increases since $|\inspace|$
will always grow as $|\mathcal{C}|^T$, while $|\mathcal{X}_+|$ will typically grow at
a lower rate.

We employ two approaches for artificially constructing balanced sets that permit
learning these models in far fewer samples than $|\mathcal{C}|^T$. In settings
where we do not have a corpus of known valid sequences, Bayesian active learning
can automatically construct the training set $\mathcal{D}$. This method works
by iteratively selecting sequences in $\inspace$ that are maximally informative
about the model parameters $w$ given the data collected so far
\citep{mackay1992information}. When we do have a set of known valid sequences,
we use these to seed a process for generating balanced sets by applying
random perturbations to valid sequences.

\subsection{Active learning}
\label{sec:active}
Let $x_{1:T}$ denote an arbitrary sequence and let $y$ be the unknown binary label
indicating whether $x_{1:T}$ is valid or not. Our model's predictive
distribution for $y$, that is, $p(y|x_{1:T},w)$ is given by (\ref{eq:predictions}).
The amount of information on $w$ that we expect to gain by
labeling and adding $x_{1:T}$ to $\mathcal{D}$ can be measured
in terms of the expected reduction in the entropy of the posterior distribution
$p(w|\mathcal{D})$. That is,
\begin{align}
  \alpha(x_{1:T}) = \text{H}[p(w |\data)] - \mathbb{E}_{p(y|x_{1:T},w)} \text{H}[{p(w | \data \cup (x_{1:T}, y)}]\,,
  \label{eq:info-gain}
\end{align}
where $\text{H}(\cdot)$ computes the entropy of a distribution. This formulation
of the entropy-based active learning criterion is, however, difficult to
approximate, because it requires us to condition on $x_{1:T}$ -- effectively . To obtain a simpler expression we follow
\citet{houlsby2011bayesian} and note that $\alpha(x_{1:T})$ is equal to the
mutual information between $y$ and $w$ given $x_{1:T}$ and $\mathcal{D}$
\begin{align}
\alpha(x_{1:T}) =
\text{H}\{\mathbb{E}_{p(w | \data)} [ p(y|x_{1:T},w)]\} - \mathbb{E}_{p(w|\data)}\{ \text{H}[p(y|x_{1:T},w)]\}\,,
\label{eq:mutual-info}
\end{align}
which is easier to work with as the required entropy is now that of Bernoulli
predictive distributions, an analytic quantity. Let $\mathcal{B}(p)$
denote a Bernoulli distribution with probability $p$, and with probability mass
$p^z(1-p)^{1-z}$ for values $z \in \{0, 1\}$. The entropy of $\mathcal{B}(p)$
can be easily obtained as
\begin{align}
\text{H}[\mathcal{B}(p)] = -p \log p - (1-p) \log(1-p)\equiv g(p)\,.\label{eq:bernoulli-entropy}
\end{align}
The expectation with respect to $p(w | \data)$ can be easily approximated by
Monte Carlo. We could attempt to sequentially construct $\mathcal{D}$ by optimising
(\ref{eq:mutual-info}). However, this optimisation process would still be
difficult, as it would require evaluating $\alpha(x_{1:T})$ exhaustively on all
the elements of $\inspace$. To avoid this, we follow a greedy approach and
construct our informative sequence in a sequential manner. In particular, at
each time step $t=1,\dots,T$, we select $x_t$ by optimising the mutual
information between $w$ and $Q^\star(x_{<t}, x_t)$, where $x_{<t}$ denotes here
the prefix already selected at previous steps of the optimisation process. This
mutual information quantity is denoted by $\alpha(x_t | x_{<t})$ and its
expression is given by
\begin{align}
\label{eq:1}
\alpha(x_t | x_{<t}) = \text{H}\{\mathbb{E}_{p(w | \data)} [\mathcal{B}(\out)]\} -
\mathbb{E}_{p(w|\data)}\{ \text{H}[\mathcal{B}(\out)] \}.
\end{align}
The generation of an informative sequence can then be performed
efficiently by sequentially optimising (\ref{eq:1}), an operation that requires
only $|\mathcal{C}|\times T$ evaluations of $\alpha(x_t | x_{<t})$.

To obtain an approximation to (\ref{eq:1}), we first approximate the posterior
distribution $p(w|\mathcal{D})$ with $q(w)$ and then estimate the expectations
in (\ref{eq:1}) by Monte Carlo using $K$ samples drawn from $q(w)$. The
resulting estimator is given by
\begin{align}
\hat \alpha(x_t \mid x_{<t}) = g\bigg[ \frac{1}{K} \sum_{k=1}^K \mathrm{y}(x_t | x_{<t}, w_k)\bigg] - \frac{1}{K} \sum_{k=1}^K g\left[ \mathrm{y}(x_t | x_{<t}, w_k) \right]\,,
\end{align}
where $w_1,\ldots,w_K \sim q(w)$ and $g(\cdot)$ is defined in
(\ref{eq:bernoulli-entropy}). The nonlinearity of $g(\cdot)$ means that our
Monte Carlo approximation is biased, but still consistent. We found that
reasonable estimates can be obtained even for small $K$. In our experiments we
use $K = 16$.

The iterative procedure just described is designed to produce a single
informative sequence. In practice, we would like to generate a batch of
informative and diverse sequences. The reason for this is that, when training
neural networks, processing a batch of data is computationally more efficient
than individually processing multiple data points. To construct a batch with $L$
informative sequences, we propose to repeat the previous iterative procedure $L$
times. To introduce diversity in the batch-generation process, we ``soften'' the
greedy maximisation operation at each step by injecting a small amount of noise
in the evaluation of the objective function \citep{finkel2006solving}. Besides
introducing diversity, this can also lead to better overall solutions than those
produced by the noiseless greedy approach \citep{cho2016noisy}. We introduce
noise into the greedy selection process by sampling from
\begin{align}
p(x_t| x_{<t},\theta) = \frac{\exp\{\alpha(x_t |x_{<t})/\theta\}}{\sum_{x_t' \in \mathcal{C}}  \exp\{\alpha(x_t' | x_{<t})/\theta\}}\,
\end{align}
for each $t=1,\dots,T$, which is a Boltzmann distribution with sampling
temperature $\theta$. By adjusting this temperature parameter, we can trade off
the diversity of samples in the batch vs. their similarity.

\subsection{Data augmentation}
\label{sec:aug}
In some settings, such as the molecule domain we will consider later, we have
databases of known-valid examples (e.g.~collections of known drug-like
molecules), but rarely are sets of invalid examples available. Obtaining invalid
sequences may seem trivial, as invalid samples may be obtained by sampling
uniformly from $\mathcal{X}$, however these are almost always so far from any
valid sequence that they carry little information about the boundary of
valid and invalid sequences. Using just a known data set also carries the danger
of overfitting to the subset of $\mathcal{X}_+$ covered by the data.

We address this by perturbing sequences from a database of valid sequences, such
that approximately half of the thus generated sequences are invalid. These
perturbed sequences $x'_{1:T}$ are constructed by setting each $x'_t$ to be a
symbol selected independently from $\mathcal{C}$ with probability $\gamma$,
while remaining the original $x_t$ with probability $1-\gamma$. In expectation
this changes $\gamma T$ entries in the sequence. We choose $\gamma = 0.05$,
which results in synthetic data that is approximately 50\% valid.


%% file: sections/python.tex

We test the proposed validity checker in two environments. First, we look at fixed length
Python 3 mathematical expressions, where we derive lower bounds for the support
of our model and compare the performance of active learning with that achieved by a
simple passive approach. Secondly, we look at molecular structures encoded into
string representation, where we utilise existing molecule data sets together
with our proposed data augmentation method to learn the rules governing molecule
string validity. We test the efficacy of our validity checker on the
downstream task of decoding valid molecules from a continuous latent representation
given by a variational autoencoder. The code to reproduce these experiments is available online\footnote{\url{https://github.com/DavidJanz/molecule_grammar_rnn}}.

\subsection{Mathematical expressions}
We illustrate the utility of the proposed validity model and sequential Bayesian
active learning in the context of Python 3 mathematical expressions. Here,
$\inspace$ consists of all length 25 sequences that can be constructed from the
alphabet of numbers and symbols shown in table \ref{tab:py-alphabet}. The
validity of any given expression is determined using the Python 3 \texttt{eval}
function: a valid expression is one that does not raise an exception when
evaluated.
\begin{table}[hb] \centering
  \caption{Python 3 expression alphabet}
  \label{tab:py-alphabet}
  \begin{tabular}{c c c c}
    \toprule
  digits & operators & comparisons & brackets \\
  \midrule
  \texttt{1234567890} & \texttt{+-*/\%!} & \texttt{=<>} & \texttt{()} \\
  \bottomrule 
\end{tabular}
\end{table}

\paragraph{Measuring model performance} Within this problem domain we do not
assume the existence of a data set of positive examples. Without a validation data
set to measure performance on, we compare the models in terms of their
capability to provide high entropy distributions over valid sequences. We define a
generative procedure to sample from the model and measure the validity and
entropy of the samples. To sample stochastically, we use a Boltzmann policy,
i.e.~a policy which samples next actions according to
\begin{align}
\pi(x_t = c | x_{<t}, w, \tau) = \frac{\exp(\mathrm{y}( c | x_{<t}, w)/\tau)}{\sum_{j \in \mathcal{C}} \exp(\mathrm{y}( j | x_{<t}, w)/\tau)}
\end{align}
where $\tau$ is a temperature constant that governs the trade-off between
exploration and exploitation. Note that this is not the same as the Boltzmann
distribution used as a proposal generation scheme during active learning, which
was defined not on $Q$-function values but rather on the estimated mutual
information.

We obtain samples $\set{x^{(1)},\dots,x^{(N)}}_{\tau_i}$ for a range of
temperatures $\tau_i$ and compute the validity fraction and entropy of each set
of samples. These points now plot a curve of the trade-off between validity and
entropy that a given model provides. Without a preferred level of sequence
validity, the area under this validity-entropy curve (V-H AUC) can be utilised
as a metric of model quality. To provide some context for the entropy values, we
estimate an information theoretic lower bound for the fraction of the set
$\inspace_{+}$ that our model is able to generate. This translates to upper
bounding the false negative rate for our model.

\begin{figure}[t]
  \centering
 \begin{subfigure}{0.49\textwidth}
    \input{figs/auc_time_expr.pgf}
  \end{subfigure}
  \begin{subfigure}{0.49\textwidth}
    \input{figs/e25_tradeoff.pgf}
  \end{subfigure}
  \caption{Experiments with length 25 Python expressions. (Left) Area under
    validity-entropy curve as training progresses, 10-90 percentiles shaded. Active learning converges
    faster and reaches a higher maximum. (Right) Entropy versus validity for
    median active and median passive model after $200$k training sequences. Both
    models have learnt a high entropy distribution over valid sequences.}
  \label{fig:expr-entropy}
\end{figure}
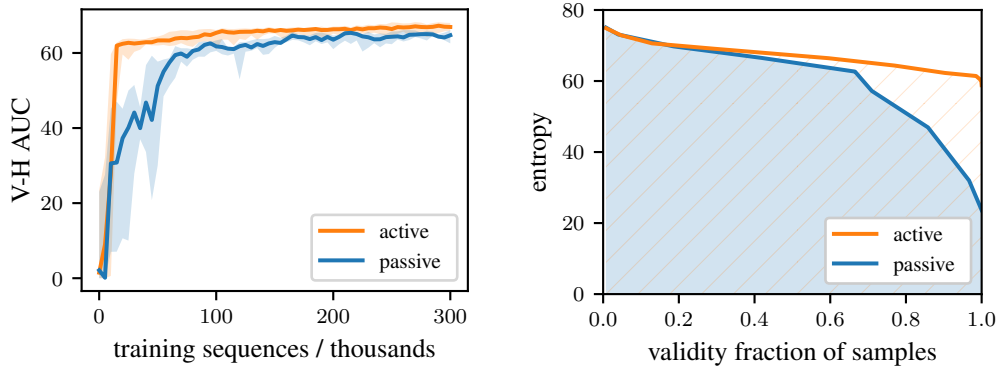

\paragraph{Experimental setup and results} We train two models using our
proposed Q-function method: \textit{passive}, where training sequences are
sampled from a uniform distribution over $\inspace$, and \textit{active}, where
we use the procedure described in section \ref{sec:active} to select training
sequences. The two models are otherwise identical.

Both trained models give a diverse output distribution over valid sequences
(figure \ref{fig:expr-entropy}). However, as expected, we find that the
\textit{active} method is able to learn a model of sequence validity much more
rapidly than sampling uniformly from $\inspace$, and the corresponding converged
model is capable of generating many more distinct valid sequences than that
trained using the \textit{passive} method. In table \ref{tab:expr-coverage} we
present lower bounds on the support of the two respective models. The details of
how this lower bound is computed can be found in appendix
\ref{appendix:coverage}. Note that the overhead of the active learning data
generating procedure is minimal: processing 10,000 takes 31s with
\textit{passive} versus 37s with \textit{active}.

\begin{table}
  \centering
  \caption{Estimated lower bound of coverage $N$ for \textit{passive} and
    \textit{active} models, defined as the size of the set of Python expressions
    on which the respective model places positive probability mass. Evaluation
    is on models trained until convergence ($800,000$ training points, beyond
    the scope of figure \ref{fig:expr-entropy}). The lower bound estimation
    method is detailed in Appendix~\ref{appendix:coverage}.}
  \begin{tabular}[h]{c  c c  c c }
    \toprule
    \multirow{2}{*}{temperature $\tau$}& \multicolumn{2}{c }{passive model} & \multicolumn{2}{c}{active model} \\
    & validity & $ N $ & validity & $ N $ \\
    \midrule
    0.100 & 0.850 & $9.7\times 10^{27}$ & 0.841 & $8.2\times 10^{28}$ \\
    0.025 & 0.969 & $2.9\times 10^{25}$ & 0.995 & $4.3\times 10^{27}$ \\
    0.005 & 1.000 & $1.1\times 10^{22}$ & 1.000 & $1.3\times 10^{27}$ \\
    \bottomrule
  \end{tabular}
  \label{tab:expr-coverage}
\end{table}


%% file: figs/auc_time_expr.pgf
\begingroup%
\makeatletter%
\begin{pgfpicture}%
\pgfpathrectangle{\pgfpointorigin}{\pgfqpoint{2.750000in}{2.200000in}}%
\pgfusepath{use as bounding box, clip}%
\begin{pgfscope}%
\pgfsetbuttcap%
\pgfsetmiterjoin%
\definecolor{currentfill}{rgb}{1.000000,1.000000,1.000000}%
\pgfsetfillcolor{currentfill}%
\pgfsetlinewidth{0.000000pt}%
\definecolor{currentstroke}{rgb}{1.000000,1.000000,1.000000}%
\pgfsetstrokecolor{currentstroke}%
\pgfsetdash{}{0pt}%
\pgfpathmoveto{\pgfqpoint{0.000000in}{0.000000in}}%
\pgfpathlineto{\pgfqpoint{2.750000in}{0.000000in}}%
\pgfpathlineto{\pgfqpoint{2.750000in}{2.200000in}}%
\pgfpathlineto{\pgfqpoint{0.000000in}{2.200000in}}%
\pgfpathclose%
\pgfusepath{fill}%
\end{pgfscope}%
\begin{pgfscope}%
\pgfsetbuttcap%
\pgfsetmiterjoin%
\definecolor{currentfill}{rgb}{1.000000,1.000000,1.000000}%
\pgfsetfillcolor{currentfill}%
\pgfsetlinewidth{0.000000pt}%
\definecolor{currentstroke}{rgb}{0.000000,0.000000,0.000000}%
\pgfsetstrokecolor{currentstroke}%
\pgfsetstrokeopacity{0.000000}%
\pgfsetdash{}{0pt}%
\pgfpathmoveto{\pgfqpoint{0.543490in}{0.539757in}}%
\pgfpathlineto{\pgfqpoint{2.565000in}{0.539757in}}%
\pgfpathlineto{\pgfqpoint{2.565000in}{2.015000in}}%
\pgfpathlineto{\pgfqpoint{0.543490in}{2.015000in}}%
\pgfpathclose%
\pgfusepath{fill}%
\end{pgfscope}%
\begin{pgfscope}%
\pgfpathrectangle{\pgfqpoint{0.543490in}{0.539757in}}{\pgfqpoint{2.021510in}{1.475243in}} %
\pgfusepath{clip}%
\pgfsetbuttcap%
\pgfsetroundjoin%
\definecolor{currentfill}{rgb}{1.000000,0.498039,0.054902}%
\pgfsetfillcolor{currentfill}%
\pgfsetfillopacity{0.200000}%
\pgfsetlinewidth{0.000000pt}%
\definecolor{currentstroke}{rgb}{1.000000,0.498039,0.054902}%
\pgfsetstrokecolor{currentstroke}%
\pgfsetstrokeopacity{0.200000}%
\pgfsetdash{}{0pt}%
\pgfpathmoveto{\pgfqpoint{0.635377in}{1.069521in}}%
\pgfpathlineto{\pgfqpoint{0.635377in}{0.606814in}}%
\pgfpathlineto{\pgfqpoint{0.666006in}{0.606814in}}%
\pgfpathlineto{\pgfqpoint{0.696635in}{0.618860in}}%
\pgfpathlineto{\pgfqpoint{0.727264in}{1.525037in}}%
\pgfpathlineto{\pgfqpoint{0.757893in}{1.778463in}}%
\pgfpathlineto{\pgfqpoint{0.788522in}{1.751094in}}%
\pgfpathlineto{\pgfqpoint{0.819151in}{1.792500in}}%
\pgfpathlineto{\pgfqpoint{0.849780in}{1.803662in}}%
\pgfpathlineto{\pgfqpoint{0.880408in}{1.831574in}}%
\pgfpathlineto{\pgfqpoint{0.911037in}{1.826583in}}%
\pgfpathlineto{\pgfqpoint{0.941666in}{1.825648in}}%
\pgfpathlineto{\pgfqpoint{0.972295in}{1.828812in}}%
\pgfpathlineto{\pgfqpoint{1.002924in}{1.822828in}}%
\pgfpathlineto{\pgfqpoint{1.033553in}{1.826503in}}%
\pgfpathlineto{\pgfqpoint{1.064182in}{1.837808in}}%
\pgfpathlineto{\pgfqpoint{1.094811in}{1.825152in}}%
\pgfpathlineto{\pgfqpoint{1.125440in}{1.818514in}}%
\pgfpathlineto{\pgfqpoint{1.156069in}{1.842685in}}%
\pgfpathlineto{\pgfqpoint{1.186698in}{1.848440in}}%
\pgfpathlineto{\pgfqpoint{1.217327in}{1.844808in}}%
\pgfpathlineto{\pgfqpoint{1.247956in}{1.803797in}}%
\pgfpathlineto{\pgfqpoint{1.278585in}{1.856546in}}%
\pgfpathlineto{\pgfqpoint{1.309214in}{1.856044in}}%
\pgfpathlineto{\pgfqpoint{1.339842in}{1.826232in}}%
\pgfpathlineto{\pgfqpoint{1.370471in}{1.860455in}}%
\pgfpathlineto{\pgfqpoint{1.401100in}{1.854720in}}%
\pgfpathlineto{\pgfqpoint{1.431729in}{1.878737in}}%
\pgfpathlineto{\pgfqpoint{1.462358in}{1.855744in}}%
\pgfpathlineto{\pgfqpoint{1.492987in}{1.878131in}}%
\pgfpathlineto{\pgfqpoint{1.523616in}{1.889296in}}%
\pgfpathlineto{\pgfqpoint{1.554245in}{1.889103in}}%
\pgfpathlineto{\pgfqpoint{1.584874in}{1.869957in}}%
\pgfpathlineto{\pgfqpoint{1.615503in}{1.846016in}}%
\pgfpathlineto{\pgfqpoint{1.646132in}{1.880799in}}%
\pgfpathlineto{\pgfqpoint{1.676761in}{1.886306in}}%
\pgfpathlineto{\pgfqpoint{1.707390in}{1.880234in}}%
\pgfpathlineto{\pgfqpoint{1.738019in}{1.879025in}}%
\pgfpathlineto{\pgfqpoint{1.768648in}{1.879548in}}%
\pgfpathlineto{\pgfqpoint{1.799277in}{1.870839in}}%
\pgfpathlineto{\pgfqpoint{1.829905in}{1.869364in}}%
\pgfpathlineto{\pgfqpoint{1.860534in}{1.887977in}}%
\pgfpathlineto{\pgfqpoint{1.891163in}{1.878884in}}%
\pgfpathlineto{\pgfqpoint{1.921792in}{1.878420in}}%
\pgfpathlineto{\pgfqpoint{1.952421in}{1.880047in}}%
\pgfpathlineto{\pgfqpoint{1.983050in}{1.886880in}}%
\pgfpathlineto{\pgfqpoint{2.013679in}{1.888281in}}%
\pgfpathlineto{\pgfqpoint{2.044308in}{1.889210in}}%
\pgfpathlineto{\pgfqpoint{2.074937in}{1.892771in}}%
\pgfpathlineto{\pgfqpoint{2.105566in}{1.884934in}}%
\pgfpathlineto{\pgfqpoint{2.136195in}{1.877653in}}%
\pgfpathlineto{\pgfqpoint{2.166824in}{1.876941in}}%
\pgfpathlineto{\pgfqpoint{2.197453in}{1.887531in}}%
\pgfpathlineto{\pgfqpoint{2.228082in}{1.898519in}}%
\pgfpathlineto{\pgfqpoint{2.258711in}{1.874239in}}%
\pgfpathlineto{\pgfqpoint{2.289340in}{1.895232in}}%
\pgfpathlineto{\pgfqpoint{2.319968in}{1.886019in}}%
\pgfpathlineto{\pgfqpoint{2.350597in}{1.870381in}}%
\pgfpathlineto{\pgfqpoint{2.381226in}{1.883623in}}%
\pgfpathlineto{\pgfqpoint{2.411855in}{1.891190in}}%
\pgfpathlineto{\pgfqpoint{2.442484in}{1.900073in}}%
\pgfpathlineto{\pgfqpoint{2.473113in}{1.903397in}}%
\pgfpathlineto{\pgfqpoint{2.473113in}{1.943978in}}%
\pgfpathlineto{\pgfqpoint{2.473113in}{1.943978in}}%
\pgfpathlineto{\pgfqpoint{2.442484in}{1.945089in}}%
\pgfpathlineto{\pgfqpoint{2.411855in}{1.947642in}}%
\pgfpathlineto{\pgfqpoint{2.381226in}{1.943289in}}%
\pgfpathlineto{\pgfqpoint{2.350597in}{1.936372in}}%
\pgfpathlineto{\pgfqpoint{2.319968in}{1.945569in}}%
\pgfpathlineto{\pgfqpoint{2.289340in}{1.947943in}}%
\pgfpathlineto{\pgfqpoint{2.258711in}{1.942276in}}%
\pgfpathlineto{\pgfqpoint{2.228082in}{1.941354in}}%
\pgfpathlineto{\pgfqpoint{2.197453in}{1.945039in}}%
\pgfpathlineto{\pgfqpoint{2.166824in}{1.946977in}}%
\pgfpathlineto{\pgfqpoint{2.136195in}{1.936942in}}%
\pgfpathlineto{\pgfqpoint{2.105566in}{1.938703in}}%
\pgfpathlineto{\pgfqpoint{2.074937in}{1.931783in}}%
\pgfpathlineto{\pgfqpoint{2.044308in}{1.934665in}}%
\pgfpathlineto{\pgfqpoint{2.013679in}{1.927775in}}%
\pgfpathlineto{\pgfqpoint{1.983050in}{1.930869in}}%
\pgfpathlineto{\pgfqpoint{1.952421in}{1.921012in}}%
\pgfpathlineto{\pgfqpoint{1.921792in}{1.928781in}}%
\pgfpathlineto{\pgfqpoint{1.891163in}{1.924884in}}%
\pgfpathlineto{\pgfqpoint{1.860534in}{1.925030in}}%
\pgfpathlineto{\pgfqpoint{1.829905in}{1.918590in}}%
\pgfpathlineto{\pgfqpoint{1.799277in}{1.924806in}}%
\pgfpathlineto{\pgfqpoint{1.768648in}{1.919771in}}%
\pgfpathlineto{\pgfqpoint{1.738019in}{1.920435in}}%
\pgfpathlineto{\pgfqpoint{1.707390in}{1.916107in}}%
\pgfpathlineto{\pgfqpoint{1.676761in}{1.920529in}}%
\pgfpathlineto{\pgfqpoint{1.646132in}{1.919940in}}%
\pgfpathlineto{\pgfqpoint{1.615503in}{1.917587in}}%
\pgfpathlineto{\pgfqpoint{1.584874in}{1.916116in}}%
\pgfpathlineto{\pgfqpoint{1.554245in}{1.915606in}}%
\pgfpathlineto{\pgfqpoint{1.523616in}{1.922495in}}%
\pgfpathlineto{\pgfqpoint{1.492987in}{1.914926in}}%
\pgfpathlineto{\pgfqpoint{1.462358in}{1.918451in}}%
\pgfpathlineto{\pgfqpoint{1.431729in}{1.912873in}}%
\pgfpathlineto{\pgfqpoint{1.401100in}{1.915284in}}%
\pgfpathlineto{\pgfqpoint{1.370471in}{1.913994in}}%
\pgfpathlineto{\pgfqpoint{1.339842in}{1.911908in}}%
\pgfpathlineto{\pgfqpoint{1.309214in}{1.912645in}}%
\pgfpathlineto{\pgfqpoint{1.278585in}{1.916296in}}%
\pgfpathlineto{\pgfqpoint{1.247956in}{1.906456in}}%
\pgfpathlineto{\pgfqpoint{1.217327in}{1.912420in}}%
\pgfpathlineto{\pgfqpoint{1.186698in}{1.911652in}}%
\pgfpathlineto{\pgfqpoint{1.156069in}{1.909469in}}%
\pgfpathlineto{\pgfqpoint{1.125440in}{1.901065in}}%
\pgfpathlineto{\pgfqpoint{1.094811in}{1.905355in}}%
\pgfpathlineto{\pgfqpoint{1.064182in}{1.905797in}}%
\pgfpathlineto{\pgfqpoint{1.033553in}{1.900551in}}%
\pgfpathlineto{\pgfqpoint{1.002924in}{1.897601in}}%
\pgfpathlineto{\pgfqpoint{0.972295in}{1.881608in}}%
\pgfpathlineto{\pgfqpoint{0.941666in}{1.872089in}}%
\pgfpathlineto{\pgfqpoint{0.911037in}{1.869933in}}%
\pgfpathlineto{\pgfqpoint{0.880408in}{1.860277in}}%
\pgfpathlineto{\pgfqpoint{0.849780in}{1.857122in}}%
\pgfpathlineto{\pgfqpoint{0.819151in}{1.860790in}}%
\pgfpathlineto{\pgfqpoint{0.788522in}{1.861843in}}%
\pgfpathlineto{\pgfqpoint{0.757893in}{1.858792in}}%
\pgfpathlineto{\pgfqpoint{0.727264in}{1.840535in}}%
\pgfpathlineto{\pgfqpoint{0.696635in}{1.780594in}}%
\pgfpathlineto{\pgfqpoint{0.666006in}{1.239127in}}%
\pgfpathlineto{\pgfqpoint{0.635377in}{1.069521in}}%
\pgfpathclose%
\pgfusepath{fill}%
\end{pgfscope}%
\begin{pgfscope}%
\pgfpathrectangle{\pgfqpoint{0.543490in}{0.539757in}}{\pgfqpoint{2.021510in}{1.475243in}} %
\pgfusepath{clip}%
\pgfsetbuttcap%
\pgfsetroundjoin%
\definecolor{currentfill}{rgb}{0.121569,0.466667,0.705882}%
\pgfsetfillcolor{currentfill}%
\pgfsetfillopacity{0.200000}%
\pgfsetlinewidth{0.000000pt}%
\definecolor{currentstroke}{rgb}{0.121569,0.466667,0.705882}%
\pgfsetstrokecolor{currentstroke}%
\pgfsetstrokeopacity{0.200000}%
\pgfsetdash{}{0pt}%
\pgfpathmoveto{\pgfqpoint{0.635377in}{1.059187in}}%
\pgfpathlineto{\pgfqpoint{0.635377in}{0.606814in}}%
\pgfpathlineto{\pgfqpoint{0.666006in}{0.606999in}}%
\pgfpathlineto{\pgfqpoint{0.696635in}{0.745006in}}%
\pgfpathlineto{\pgfqpoint{0.727264in}{0.745726in}}%
\pgfpathlineto{\pgfqpoint{0.757893in}{0.815912in}}%
\pgfpathlineto{\pgfqpoint{0.788522in}{0.803718in}}%
\pgfpathlineto{\pgfqpoint{0.819151in}{1.155005in}}%
\pgfpathlineto{\pgfqpoint{0.849780in}{1.334786in}}%
\pgfpathlineto{\pgfqpoint{0.880408in}{1.267911in}}%
\pgfpathlineto{\pgfqpoint{0.911037in}{1.015358in}}%
\pgfpathlineto{\pgfqpoint{0.941666in}{1.197055in}}%
\pgfpathlineto{\pgfqpoint{0.972295in}{1.559544in}}%
\pgfpathlineto{\pgfqpoint{1.002924in}{1.613201in}}%
\pgfpathlineto{\pgfqpoint{1.033553in}{1.656511in}}%
\pgfpathlineto{\pgfqpoint{1.064182in}{1.740877in}}%
\pgfpathlineto{\pgfqpoint{1.094811in}{1.714573in}}%
\pgfpathlineto{\pgfqpoint{1.125440in}{1.751132in}}%
\pgfpathlineto{\pgfqpoint{1.156069in}{1.767424in}}%
\pgfpathlineto{\pgfqpoint{1.186698in}{1.766823in}}%
\pgfpathlineto{\pgfqpoint{1.217327in}{1.789806in}}%
\pgfpathlineto{\pgfqpoint{1.247956in}{1.797405in}}%
\pgfpathlineto{\pgfqpoint{1.278585in}{1.779763in}}%
\pgfpathlineto{\pgfqpoint{1.309214in}{1.773120in}}%
\pgfpathlineto{\pgfqpoint{1.339842in}{1.782979in}}%
\pgfpathlineto{\pgfqpoint{1.370471in}{1.648227in}}%
\pgfpathlineto{\pgfqpoint{1.401100in}{1.793281in}}%
\pgfpathlineto{\pgfqpoint{1.431729in}{1.798014in}}%
\pgfpathlineto{\pgfqpoint{1.462358in}{1.812423in}}%
\pgfpathlineto{\pgfqpoint{1.492987in}{1.776972in}}%
\pgfpathlineto{\pgfqpoint{1.523616in}{1.781267in}}%
\pgfpathlineto{\pgfqpoint{1.554245in}{1.813286in}}%
\pgfpathlineto{\pgfqpoint{1.584874in}{1.836463in}}%
\pgfpathlineto{\pgfqpoint{1.615503in}{1.842987in}}%
\pgfpathlineto{\pgfqpoint{1.646132in}{1.854897in}}%
\pgfpathlineto{\pgfqpoint{1.676761in}{1.853108in}}%
\pgfpathlineto{\pgfqpoint{1.707390in}{1.841497in}}%
\pgfpathlineto{\pgfqpoint{1.738019in}{1.826864in}}%
\pgfpathlineto{\pgfqpoint{1.768648in}{1.847789in}}%
\pgfpathlineto{\pgfqpoint{1.799277in}{1.833224in}}%
\pgfpathlineto{\pgfqpoint{1.829905in}{1.842453in}}%
\pgfpathlineto{\pgfqpoint{1.860534in}{1.806603in}}%
\pgfpathlineto{\pgfqpoint{1.891163in}{1.831744in}}%
\pgfpathlineto{\pgfqpoint{1.921792in}{1.850939in}}%
\pgfpathlineto{\pgfqpoint{1.952421in}{1.833952in}}%
\pgfpathlineto{\pgfqpoint{1.983050in}{1.851866in}}%
\pgfpathlineto{\pgfqpoint{2.013679in}{1.793981in}}%
\pgfpathlineto{\pgfqpoint{2.044308in}{1.814016in}}%
\pgfpathlineto{\pgfqpoint{2.074937in}{1.837783in}}%
\pgfpathlineto{\pgfqpoint{2.105566in}{1.816298in}}%
\pgfpathlineto{\pgfqpoint{2.136195in}{1.836429in}}%
\pgfpathlineto{\pgfqpoint{2.166824in}{1.832528in}}%
\pgfpathlineto{\pgfqpoint{2.197453in}{1.856956in}}%
\pgfpathlineto{\pgfqpoint{2.228082in}{1.846746in}}%
\pgfpathlineto{\pgfqpoint{2.258711in}{1.854381in}}%
\pgfpathlineto{\pgfqpoint{2.289340in}{1.870178in}}%
\pgfpathlineto{\pgfqpoint{2.319968in}{1.862433in}}%
\pgfpathlineto{\pgfqpoint{2.350597in}{1.851866in}}%
\pgfpathlineto{\pgfqpoint{2.381226in}{1.877350in}}%
\pgfpathlineto{\pgfqpoint{2.411855in}{1.867386in}}%
\pgfpathlineto{\pgfqpoint{2.442484in}{1.852862in}}%
\pgfpathlineto{\pgfqpoint{2.473113in}{1.836732in}}%
\pgfpathlineto{\pgfqpoint{2.473113in}{1.906016in}}%
\pgfpathlineto{\pgfqpoint{2.473113in}{1.906016in}}%
\pgfpathlineto{\pgfqpoint{2.442484in}{1.893516in}}%
\pgfpathlineto{\pgfqpoint{2.411855in}{1.884829in}}%
\pgfpathlineto{\pgfqpoint{2.381226in}{1.898049in}}%
\pgfpathlineto{\pgfqpoint{2.350597in}{1.903678in}}%
\pgfpathlineto{\pgfqpoint{2.319968in}{1.893508in}}%
\pgfpathlineto{\pgfqpoint{2.289340in}{1.904840in}}%
\pgfpathlineto{\pgfqpoint{2.258711in}{1.906452in}}%
\pgfpathlineto{\pgfqpoint{2.228082in}{1.906267in}}%
\pgfpathlineto{\pgfqpoint{2.197453in}{1.899990in}}%
\pgfpathlineto{\pgfqpoint{2.166824in}{1.899297in}}%
\pgfpathlineto{\pgfqpoint{2.136195in}{1.886663in}}%
\pgfpathlineto{\pgfqpoint{2.105566in}{1.889135in}}%
\pgfpathlineto{\pgfqpoint{2.074937in}{1.883413in}}%
\pgfpathlineto{\pgfqpoint{2.044308in}{1.897054in}}%
\pgfpathlineto{\pgfqpoint{2.013679in}{1.885369in}}%
\pgfpathlineto{\pgfqpoint{1.983050in}{1.894925in}}%
\pgfpathlineto{\pgfqpoint{1.952421in}{1.898351in}}%
\pgfpathlineto{\pgfqpoint{1.921792in}{1.899093in}}%
\pgfpathlineto{\pgfqpoint{1.891163in}{1.881363in}}%
\pgfpathlineto{\pgfqpoint{1.860534in}{1.878303in}}%
\pgfpathlineto{\pgfqpoint{1.829905in}{1.886324in}}%
\pgfpathlineto{\pgfqpoint{1.799277in}{1.893088in}}%
\pgfpathlineto{\pgfqpoint{1.768648in}{1.891306in}}%
\pgfpathlineto{\pgfqpoint{1.738019in}{1.879501in}}%
\pgfpathlineto{\pgfqpoint{1.707390in}{1.881444in}}%
\pgfpathlineto{\pgfqpoint{1.676761in}{1.885202in}}%
\pgfpathlineto{\pgfqpoint{1.646132in}{1.899616in}}%
\pgfpathlineto{\pgfqpoint{1.615503in}{1.877350in}}%
\pgfpathlineto{\pgfqpoint{1.584874in}{1.866240in}}%
\pgfpathlineto{\pgfqpoint{1.554245in}{1.876014in}}%
\pgfpathlineto{\pgfqpoint{1.523616in}{1.862395in}}%
\pgfpathlineto{\pgfqpoint{1.492987in}{1.859087in}}%
\pgfpathlineto{\pgfqpoint{1.462358in}{1.860439in}}%
\pgfpathlineto{\pgfqpoint{1.431729in}{1.852451in}}%
\pgfpathlineto{\pgfqpoint{1.401100in}{1.843349in}}%
\pgfpathlineto{\pgfqpoint{1.370471in}{1.842616in}}%
\pgfpathlineto{\pgfqpoint{1.339842in}{1.820512in}}%
\pgfpathlineto{\pgfqpoint{1.309214in}{1.813903in}}%
\pgfpathlineto{\pgfqpoint{1.278585in}{1.859758in}}%
\pgfpathlineto{\pgfqpoint{1.247956in}{1.852464in}}%
\pgfpathlineto{\pgfqpoint{1.217327in}{1.844074in}}%
\pgfpathlineto{\pgfqpoint{1.186698in}{1.848353in}}%
\pgfpathlineto{\pgfqpoint{1.156069in}{1.842531in}}%
\pgfpathlineto{\pgfqpoint{1.125440in}{1.805735in}}%
\pgfpathlineto{\pgfqpoint{1.094811in}{1.811074in}}%
\pgfpathlineto{\pgfqpoint{1.064182in}{1.811484in}}%
\pgfpathlineto{\pgfqpoint{1.033553in}{1.822324in}}%
\pgfpathlineto{\pgfqpoint{1.002924in}{1.759492in}}%
\pgfpathlineto{\pgfqpoint{0.972295in}{1.752577in}}%
\pgfpathlineto{\pgfqpoint{0.941666in}{1.741267in}}%
\pgfpathlineto{\pgfqpoint{0.911037in}{1.772402in}}%
\pgfpathlineto{\pgfqpoint{0.880408in}{1.723190in}}%
\pgfpathlineto{\pgfqpoint{0.849780in}{1.502025in}}%
\pgfpathlineto{\pgfqpoint{0.819151in}{1.619237in}}%
\pgfpathlineto{\pgfqpoint{0.788522in}{1.589956in}}%
\pgfpathlineto{\pgfqpoint{0.757893in}{1.494579in}}%
\pgfpathlineto{\pgfqpoint{0.727264in}{1.531479in}}%
\pgfpathlineto{\pgfqpoint{0.696635in}{1.495572in}}%
\pgfpathlineto{\pgfqpoint{0.666006in}{1.148804in}}%
\pgfpathlineto{\pgfqpoint{0.635377in}{1.059187in}}%
\pgfpathclose%
\pgfusepath{fill}%
\end{pgfscope}%
\begin{pgfscope}%
\pgfsetbuttcap%
\pgfsetroundjoin%
\definecolor{currentfill}{rgb}{0.000000,0.000000,0.000000}%
\pgfsetfillcolor{currentfill}%
\pgfsetlinewidth{0.803000pt}%
\definecolor{currentstroke}{rgb}{0.000000,0.000000,0.000000}%
\pgfsetstrokecolor{currentstroke}%
\pgfsetdash{}{0pt}%
\pgfsys@defobject{currentmarker}{\pgfqpoint{0.000000in}{-0.048611in}}{\pgfqpoint{0.000000in}{0.000000in}}{%
\pgfpathmoveto{\pgfqpoint{0.000000in}{0.000000in}}%
\pgfpathlineto{\pgfqpoint{0.000000in}{-0.048611in}}%
\pgfusepath{stroke,fill}%
}%
\begin{pgfscope}%
\pgfsys@transformshift{0.635377in}{0.539757in}%
\pgfsys@useobject{currentmarker}{}%
\end{pgfscope}%
\end{pgfscope}%
\begin{pgfscope}%
\pgftext[x=0.635377in,y=0.442535in,,top]{\rmfamily\fontsize{8.000000}{9.600000}\selectfont \(\displaystyle 0\)}%
\end{pgfscope}%
\begin{pgfscope}%
\pgfsetbuttcap%
\pgfsetroundjoin%
\definecolor{currentfill}{rgb}{0.000000,0.000000,0.000000}%
\pgfsetfillcolor{currentfill}%
\pgfsetlinewidth{0.803000pt}%
\definecolor{currentstroke}{rgb}{0.000000,0.000000,0.000000}%
\pgfsetstrokecolor{currentstroke}%
\pgfsetdash{}{0pt}%
\pgfsys@defobject{currentmarker}{\pgfqpoint{0.000000in}{-0.048611in}}{\pgfqpoint{0.000000in}{0.000000in}}{%
\pgfpathmoveto{\pgfqpoint{0.000000in}{0.000000in}}%
\pgfpathlineto{\pgfqpoint{0.000000in}{-0.048611in}}%
\pgfusepath{stroke,fill}%
}%
\begin{pgfscope}%
\pgfsys@transformshift{1.247956in}{0.539757in}%
\pgfsys@useobject{currentmarker}{}%
\end{pgfscope}%
\end{pgfscope}%
\begin{pgfscope}%
\pgftext[x=1.247956in,y=0.442535in,,top]{\rmfamily\fontsize{8.000000}{9.600000}\selectfont \(\displaystyle 100\)}%
\end{pgfscope}%
\begin{pgfscope}%
\pgfsetbuttcap%
\pgfsetroundjoin%
\definecolor{currentfill}{rgb}{0.000000,0.000000,0.000000}%
\pgfsetfillcolor{currentfill}%
\pgfsetlinewidth{0.803000pt}%
\definecolor{currentstroke}{rgb}{0.000000,0.000000,0.000000}%
\pgfsetstrokecolor{currentstroke}%
\pgfsetdash{}{0pt}%
\pgfsys@defobject{currentmarker}{\pgfqpoint{0.000000in}{-0.048611in}}{\pgfqpoint{0.000000in}{0.000000in}}{%
\pgfpathmoveto{\pgfqpoint{0.000000in}{0.000000in}}%
\pgfpathlineto{\pgfqpoint{0.000000in}{-0.048611in}}%
\pgfusepath{stroke,fill}%
}%
\begin{pgfscope}%
\pgfsys@transformshift{1.860534in}{0.539757in}%
\pgfsys@useobject{currentmarker}{}%
\end{pgfscope}%
\end{pgfscope}%
\begin{pgfscope}%
\pgftext[x=1.860534in,y=0.442535in,,top]{\rmfamily\fontsize{8.000000}{9.600000}\selectfont \(\displaystyle 200\)}%
\end{pgfscope}%
\begin{pgfscope}%
\pgfsetbuttcap%
\pgfsetroundjoin%
\definecolor{currentfill}{rgb}{0.000000,0.000000,0.000000}%
\pgfsetfillcolor{currentfill}%
\pgfsetlinewidth{0.803000pt}%
\definecolor{currentstroke}{rgb}{0.000000,0.000000,0.000000}%
\pgfsetstrokecolor{currentstroke}%
\pgfsetdash{}{0pt}%
\pgfsys@defobject{currentmarker}{\pgfqpoint{0.000000in}{-0.048611in}}{\pgfqpoint{0.000000in}{0.000000in}}{%
\pgfpathmoveto{\pgfqpoint{0.000000in}{0.000000in}}%
\pgfpathlineto{\pgfqpoint{0.000000in}{-0.048611in}}%
\pgfusepath{stroke,fill}%
}%
\begin{pgfscope}%
\pgfsys@transformshift{2.473113in}{0.539757in}%
\pgfsys@useobject{currentmarker}{}%
\end{pgfscope}%
\end{pgfscope}%
\begin{pgfscope}%
\pgftext[x=2.473113in,y=0.442535in,,top]{\rmfamily\fontsize{8.000000}{9.600000}\selectfont \(\displaystyle 300\)}%
\end{pgfscope}%
\begin{pgfscope}%
\pgftext[x=1.554245in,y=0.288855in,,top]{\rmfamily\fontsize{10.000000}{12.000000}\selectfont training sequences / thousands}%
\end{pgfscope}%
\begin{pgfscope}%
\pgfsetbuttcap%
\pgfsetroundjoin%
\definecolor{currentfill}{rgb}{0.000000,0.000000,0.000000}%
\pgfsetfillcolor{currentfill}%
\pgfsetlinewidth{0.803000pt}%
\definecolor{currentstroke}{rgb}{0.000000,0.000000,0.000000}%
\pgfsetstrokecolor{currentstroke}%
\pgfsetdash{}{0pt}%
\pgfsys@defobject{currentmarker}{\pgfqpoint{-0.048611in}{0.000000in}}{\pgfqpoint{0.000000in}{0.000000in}}{%
\pgfpathmoveto{\pgfqpoint{0.000000in}{0.000000in}}%
\pgfpathlineto{\pgfqpoint{-0.048611in}{0.000000in}}%
\pgfusepath{stroke,fill}%
}%
\begin{pgfscope}%
\pgfsys@transformshift{0.543490in}{0.606814in}%
\pgfsys@useobject{currentmarker}{}%
\end{pgfscope}%
\end{pgfscope}%
\begin{pgfscope}%
\pgftext[x=0.387239in,y=0.568551in,left,base]{\rmfamily\fontsize{8.000000}{9.600000}\selectfont \(\displaystyle 0\)}%
\end{pgfscope}%
\begin{pgfscope}%
\pgfsetbuttcap%
\pgfsetroundjoin%
\definecolor{currentfill}{rgb}{0.000000,0.000000,0.000000}%
\pgfsetfillcolor{currentfill}%
\pgfsetlinewidth{0.803000pt}%
\definecolor{currentstroke}{rgb}{0.000000,0.000000,0.000000}%
\pgfsetstrokecolor{currentstroke}%
\pgfsetdash{}{0pt}%
\pgfsys@defobject{currentmarker}{\pgfqpoint{-0.048611in}{0.000000in}}{\pgfqpoint{0.000000in}{0.000000in}}{%
\pgfpathmoveto{\pgfqpoint{0.000000in}{0.000000in}}%
\pgfpathlineto{\pgfqpoint{-0.048611in}{0.000000in}}%
\pgfusepath{stroke,fill}%
}%
\begin{pgfscope}%
\pgfsys@transformshift{0.543490in}{1.000077in}%
\pgfsys@useobject{currentmarker}{}%
\end{pgfscope}%
\end{pgfscope}%
\begin{pgfscope}%
\pgftext[x=0.328211in,y=0.961815in,left,base]{\rmfamily\fontsize{8.000000}{9.600000}\selectfont \(\displaystyle 20\)}%
\end{pgfscope}%
\begin{pgfscope}%
\pgfsetbuttcap%
\pgfsetroundjoin%
\definecolor{currentfill}{rgb}{0.000000,0.000000,0.000000}%
\pgfsetfillcolor{currentfill}%
\pgfsetlinewidth{0.803000pt}%
\definecolor{currentstroke}{rgb}{0.000000,0.000000,0.000000}%
\pgfsetstrokecolor{currentstroke}%
\pgfsetdash{}{0pt}%
\pgfsys@defobject{currentmarker}{\pgfqpoint{-0.048611in}{0.000000in}}{\pgfqpoint{0.000000in}{0.000000in}}{%
\pgfpathmoveto{\pgfqpoint{0.000000in}{0.000000in}}%
\pgfpathlineto{\pgfqpoint{-0.048611in}{0.000000in}}%
\pgfusepath{stroke,fill}%
}%
\begin{pgfscope}%
\pgfsys@transformshift{0.543490in}{1.393341in}%
\pgfsys@useobject{currentmarker}{}%
\end{pgfscope}%
\end{pgfscope}%
\begin{pgfscope}%
\pgftext[x=0.328211in,y=1.355079in,left,base]{\rmfamily\fontsize{8.000000}{9.600000}\selectfont \(\displaystyle 40\)}%
\end{pgfscope}%
\begin{pgfscope}%
\pgfsetbuttcap%
\pgfsetroundjoin%
\definecolor{currentfill}{rgb}{0.000000,0.000000,0.000000}%
\pgfsetfillcolor{currentfill}%
\pgfsetlinewidth{0.803000pt}%
\definecolor{currentstroke}{rgb}{0.000000,0.000000,0.000000}%
\pgfsetstrokecolor{currentstroke}%
\pgfsetdash{}{0pt}%
\pgfsys@defobject{currentmarker}{\pgfqpoint{-0.048611in}{0.000000in}}{\pgfqpoint{0.000000in}{0.000000in}}{%
\pgfpathmoveto{\pgfqpoint{0.000000in}{0.000000in}}%
\pgfpathlineto{\pgfqpoint{-0.048611in}{0.000000in}}%
\pgfusepath{stroke,fill}%
}%
\begin{pgfscope}%
\pgfsys@transformshift{0.543490in}{1.786605in}%
\pgfsys@useobject{currentmarker}{}%
\end{pgfscope}%
\end{pgfscope}%
\begin{pgfscope}%
\pgftext[x=0.328211in,y=1.748343in,left,base]{\rmfamily\fontsize{8.000000}{9.600000}\selectfont \(\displaystyle 60\)}%
\end{pgfscope}%
\begin{pgfscope}%
\pgftext[x=0.272655in,y=1.277379in,,bottom,rotate=90.000000]{\rmfamily\fontsize{10.000000}{12.000000}\selectfont V-H AUC}%
\end{pgfscope}%
\begin{pgfscope}%
\pgfpathrectangle{\pgfqpoint{0.543490in}{0.539757in}}{\pgfqpoint{2.021510in}{1.475243in}} %
\pgfusepath{clip}%
\pgfsetrectcap%
\pgfsetroundjoin%
\pgfsetlinewidth{1.505625pt}%
\definecolor{currentstroke}{rgb}{1.000000,0.498039,0.054902}%
\pgfsetstrokecolor{currentstroke}%
\pgfsetdash{}{0pt}%
\pgfpathmoveto{\pgfqpoint{0.635377in}{0.635848in}}%
\pgfpathlineto{\pgfqpoint{0.666006in}{0.783569in}}%
\pgfpathlineto{\pgfqpoint{0.696635in}{1.154428in}}%
\pgfpathlineto{\pgfqpoint{0.727264in}{1.825059in}}%
\pgfpathlineto{\pgfqpoint{0.757893in}{1.834722in}}%
\pgfpathlineto{\pgfqpoint{0.788522in}{1.840009in}}%
\pgfpathlineto{\pgfqpoint{0.819151in}{1.836180in}}%
\pgfpathlineto{\pgfqpoint{0.849780in}{1.839410in}}%
\pgfpathlineto{\pgfqpoint{0.880408in}{1.843012in}}%
\pgfpathlineto{\pgfqpoint{0.911037in}{1.842862in}}%
\pgfpathlineto{\pgfqpoint{0.941666in}{1.852253in}}%
\pgfpathlineto{\pgfqpoint{0.972295in}{1.852479in}}%
\pgfpathlineto{\pgfqpoint{1.002924in}{1.851877in}}%
\pgfpathlineto{\pgfqpoint{1.033553in}{1.862243in}}%
\pgfpathlineto{\pgfqpoint{1.064182in}{1.865638in}}%
\pgfpathlineto{\pgfqpoint{1.094811in}{1.863835in}}%
\pgfpathlineto{\pgfqpoint{1.125440in}{1.868313in}}%
\pgfpathlineto{\pgfqpoint{1.156069in}{1.871299in}}%
\pgfpathlineto{\pgfqpoint{1.186698in}{1.884291in}}%
\pgfpathlineto{\pgfqpoint{1.217327in}{1.880559in}}%
\pgfpathlineto{\pgfqpoint{1.247956in}{1.892082in}}%
\pgfpathlineto{\pgfqpoint{1.278585in}{1.901120in}}%
\pgfpathlineto{\pgfqpoint{1.309214in}{1.895090in}}%
\pgfpathlineto{\pgfqpoint{1.339842in}{1.893006in}}%
\pgfpathlineto{\pgfqpoint{1.370471in}{1.897087in}}%
\pgfpathlineto{\pgfqpoint{1.401100in}{1.897776in}}%
\pgfpathlineto{\pgfqpoint{1.431729in}{1.897822in}}%
\pgfpathlineto{\pgfqpoint{1.462358in}{1.902775in}}%
\pgfpathlineto{\pgfqpoint{1.492987in}{1.899888in}}%
\pgfpathlineto{\pgfqpoint{1.523616in}{1.906925in}}%
\pgfpathlineto{\pgfqpoint{1.554245in}{1.902122in}}%
\pgfpathlineto{\pgfqpoint{1.584874in}{1.907079in}}%
\pgfpathlineto{\pgfqpoint{1.615503in}{1.902597in}}%
\pgfpathlineto{\pgfqpoint{1.646132in}{1.905114in}}%
\pgfpathlineto{\pgfqpoint{1.676761in}{1.908797in}}%
\pgfpathlineto{\pgfqpoint{1.707390in}{1.906892in}}%
\pgfpathlineto{\pgfqpoint{1.738019in}{1.909121in}}%
\pgfpathlineto{\pgfqpoint{1.768648in}{1.912224in}}%
\pgfpathlineto{\pgfqpoint{1.799277in}{1.906621in}}%
\pgfpathlineto{\pgfqpoint{1.829905in}{1.907360in}}%
\pgfpathlineto{\pgfqpoint{1.860534in}{1.910698in}}%
\pgfpathlineto{\pgfqpoint{1.891163in}{1.910853in}}%
\pgfpathlineto{\pgfqpoint{1.921792in}{1.913042in}}%
\pgfpathlineto{\pgfqpoint{1.952421in}{1.905698in}}%
\pgfpathlineto{\pgfqpoint{1.983050in}{1.914777in}}%
\pgfpathlineto{\pgfqpoint{2.013679in}{1.910188in}}%
\pgfpathlineto{\pgfqpoint{2.044308in}{1.916245in}}%
\pgfpathlineto{\pgfqpoint{2.074937in}{1.916131in}}%
\pgfpathlineto{\pgfqpoint{2.105566in}{1.919927in}}%
\pgfpathlineto{\pgfqpoint{2.136195in}{1.917913in}}%
\pgfpathlineto{\pgfqpoint{2.166824in}{1.911972in}}%
\pgfpathlineto{\pgfqpoint{2.197453in}{1.922161in}}%
\pgfpathlineto{\pgfqpoint{2.228082in}{1.920172in}}%
\pgfpathlineto{\pgfqpoint{2.258711in}{1.921547in}}%
\pgfpathlineto{\pgfqpoint{2.289340in}{1.925995in}}%
\pgfpathlineto{\pgfqpoint{2.319968in}{1.922728in}}%
\pgfpathlineto{\pgfqpoint{2.350597in}{1.921685in}}%
\pgfpathlineto{\pgfqpoint{2.381226in}{1.923453in}}%
\pgfpathlineto{\pgfqpoint{2.411855in}{1.929385in}}%
\pgfpathlineto{\pgfqpoint{2.442484in}{1.924118in}}%
\pgfpathlineto{\pgfqpoint{2.473113in}{1.922817in}}%
\pgfusepath{stroke}%
\end{pgfscope}%
\begin{pgfscope}%
\pgfpathrectangle{\pgfqpoint{0.543490in}{0.539757in}}{\pgfqpoint{2.021510in}{1.475243in}} %
\pgfusepath{clip}%
\pgfsetrectcap%
\pgfsetroundjoin%
\pgfsetlinewidth{1.505625pt}%
\definecolor{currentstroke}{rgb}{0.121569,0.466667,0.705882}%
\pgfsetstrokecolor{currentstroke}%
\pgfsetdash{}{0pt}%
\pgfpathmoveto{\pgfqpoint{0.635377in}{0.648447in}}%
\pgfpathlineto{\pgfqpoint{0.666006in}{0.609142in}}%
\pgfpathlineto{\pgfqpoint{0.696635in}{1.208588in}}%
\pgfpathlineto{\pgfqpoint{0.727264in}{1.212518in}}%
\pgfpathlineto{\pgfqpoint{0.757893in}{1.339957in}}%
\pgfpathlineto{\pgfqpoint{0.788522in}{1.395982in}}%
\pgfpathlineto{\pgfqpoint{0.819151in}{1.474741in}}%
\pgfpathlineto{\pgfqpoint{0.849780in}{1.392703in}}%
\pgfpathlineto{\pgfqpoint{0.880408in}{1.527002in}}%
\pgfpathlineto{\pgfqpoint{0.911037in}{1.435621in}}%
\pgfpathlineto{\pgfqpoint{0.941666in}{1.613847in}}%
\pgfpathlineto{\pgfqpoint{0.972295in}{1.688478in}}%
\pgfpathlineto{\pgfqpoint{1.002924in}{1.736755in}}%
\pgfpathlineto{\pgfqpoint{1.033553in}{1.775802in}}%
\pgfpathlineto{\pgfqpoint{1.064182in}{1.783979in}}%
\pgfpathlineto{\pgfqpoint{1.094811in}{1.766769in}}%
\pgfpathlineto{\pgfqpoint{1.125440in}{1.797421in}}%
\pgfpathlineto{\pgfqpoint{1.156069in}{1.804825in}}%
\pgfpathlineto{\pgfqpoint{1.186698in}{1.828808in}}%
\pgfpathlineto{\pgfqpoint{1.217327in}{1.837324in}}%
\pgfpathlineto{\pgfqpoint{1.247956in}{1.821132in}}%
\pgfpathlineto{\pgfqpoint{1.278585in}{1.817755in}}%
\pgfpathlineto{\pgfqpoint{1.309214in}{1.809231in}}%
\pgfpathlineto{\pgfqpoint{1.339842in}{1.806826in}}%
\pgfpathlineto{\pgfqpoint{1.370471in}{1.822080in}}%
\pgfpathlineto{\pgfqpoint{1.401100in}{1.828858in}}%
\pgfpathlineto{\pgfqpoint{1.431729in}{1.815146in}}%
\pgfpathlineto{\pgfqpoint{1.462358in}{1.833488in}}%
\pgfpathlineto{\pgfqpoint{1.492987in}{1.823397in}}%
\pgfpathlineto{\pgfqpoint{1.523616in}{1.836763in}}%
\pgfpathlineto{\pgfqpoint{1.554245in}{1.843285in}}%
\pgfpathlineto{\pgfqpoint{1.584874in}{1.839121in}}%
\pgfpathlineto{\pgfqpoint{1.615503in}{1.858318in}}%
\pgfpathlineto{\pgfqpoint{1.646132in}{1.878099in}}%
\pgfpathlineto{\pgfqpoint{1.676761in}{1.870080in}}%
\pgfpathlineto{\pgfqpoint{1.707390in}{1.868905in}}%
\pgfpathlineto{\pgfqpoint{1.738019in}{1.859276in}}%
\pgfpathlineto{\pgfqpoint{1.768648in}{1.869850in}}%
\pgfpathlineto{\pgfqpoint{1.799277in}{1.854037in}}%
\pgfpathlineto{\pgfqpoint{1.829905in}{1.870783in}}%
\pgfpathlineto{\pgfqpoint{1.860534in}{1.856024in}}%
\pgfpathlineto{\pgfqpoint{1.891163in}{1.871027in}}%
\pgfpathlineto{\pgfqpoint{1.921792in}{1.889777in}}%
\pgfpathlineto{\pgfqpoint{1.952421in}{1.891985in}}%
\pgfpathlineto{\pgfqpoint{1.983050in}{1.885884in}}%
\pgfpathlineto{\pgfqpoint{2.013679in}{1.873456in}}%
\pgfpathlineto{\pgfqpoint{2.044308in}{1.868772in}}%
\pgfpathlineto{\pgfqpoint{2.074937in}{1.857908in}}%
\pgfpathlineto{\pgfqpoint{2.105566in}{1.858412in}}%
\pgfpathlineto{\pgfqpoint{2.136195in}{1.869560in}}%
\pgfpathlineto{\pgfqpoint{2.166824in}{1.871536in}}%
\pgfpathlineto{\pgfqpoint{2.197453in}{1.873555in}}%
\pgfpathlineto{\pgfqpoint{2.228082in}{1.867923in}}%
\pgfpathlineto{\pgfqpoint{2.258711in}{1.881825in}}%
\pgfpathlineto{\pgfqpoint{2.289340in}{1.882992in}}%
\pgfpathlineto{\pgfqpoint{2.319968in}{1.885081in}}%
\pgfpathlineto{\pgfqpoint{2.350597in}{1.885437in}}%
\pgfpathlineto{\pgfqpoint{2.381226in}{1.883746in}}%
\pgfpathlineto{\pgfqpoint{2.411855in}{1.872011in}}%
\pgfpathlineto{\pgfqpoint{2.442484in}{1.867809in}}%
\pgfpathlineto{\pgfqpoint{2.473113in}{1.879451in}}%
\pgfusepath{stroke}%
\end{pgfscope}%
\begin{pgfscope}%
\pgfsetrectcap%
\pgfsetmiterjoin%
\pgfsetlinewidth{0.803000pt}%
\definecolor{currentstroke}{rgb}{0.000000,0.000000,0.000000}%
\pgfsetstrokecolor{currentstroke}%
\pgfsetdash{}{0pt}%
\pgfpathmoveto{\pgfqpoint{0.543490in}{0.539757in}}%
\pgfpathlineto{\pgfqpoint{0.543490in}{2.015000in}}%
\pgfusepath{stroke}%
\end{pgfscope}%
\begin{pgfscope}%
\pgfsetrectcap%
\pgfsetmiterjoin%
\pgfsetlinewidth{0.803000pt}%
\definecolor{currentstroke}{rgb}{0.000000,0.000000,0.000000}%
\pgfsetstrokecolor{currentstroke}%
\pgfsetdash{}{0pt}%
\pgfpathmoveto{\pgfqpoint{2.565000in}{0.539757in}}%
\pgfpathlineto{\pgfqpoint{2.565000in}{2.015000in}}%
\pgfusepath{stroke}%
\end{pgfscope}%
\begin{pgfscope}%
\pgfsetrectcap%
\pgfsetmiterjoin%
\pgfsetlinewidth{0.803000pt}%
\definecolor{currentstroke}{rgb}{0.000000,0.000000,0.000000}%
\pgfsetstrokecolor{currentstroke}%
\pgfsetdash{}{0pt}%
\pgfpathmoveto{\pgfqpoint{0.543490in}{0.539757in}}%
\pgfpathlineto{\pgfqpoint{2.565000in}{0.539757in}}%
\pgfusepath{stroke}%
\end{pgfscope}%
\begin{pgfscope}%
\pgfsetrectcap%
\pgfsetmiterjoin%
\pgfsetlinewidth{0.803000pt}%
\definecolor{currentstroke}{rgb}{0.000000,0.000000,0.000000}%
\pgfsetstrokecolor{currentstroke}%
\pgfsetdash{}{0pt}%
\pgfpathmoveto{\pgfqpoint{0.543490in}{2.015000in}}%
\pgfpathlineto{\pgfqpoint{2.565000in}{2.015000in}}%
\pgfusepath{stroke}%
\end{pgfscope}%
\begin{pgfscope}%
\pgfsetbuttcap%
\pgfsetmiterjoin%
\definecolor{currentfill}{rgb}{1.000000,1.000000,1.000000}%
\pgfsetfillcolor{currentfill}%
\pgfsetfillopacity{0.800000}%
\pgfsetlinewidth{1.003750pt}%
\definecolor{currentstroke}{rgb}{0.800000,0.800000,0.800000}%
\pgfsetstrokecolor{currentstroke}%
\pgfsetstrokeopacity{0.800000}%
\pgfsetdash{}{0pt}%
\pgfpathmoveto{\pgfqpoint{1.769718in}{0.595313in}}%
\pgfpathlineto{\pgfqpoint{2.487222in}{0.595313in}}%
\pgfpathquadraticcurveto{\pgfqpoint{2.509444in}{0.595313in}}{\pgfqpoint{2.509444in}{0.617535in}}%
\pgfpathlineto{\pgfqpoint{2.509444in}{0.916290in}}%
\pgfpathquadraticcurveto{\pgfqpoint{2.509444in}{0.938512in}}{\pgfqpoint{2.487222in}{0.938512in}}%
\pgfpathlineto{\pgfqpoint{1.769718in}{0.938512in}}%
\pgfpathquadraticcurveto{\pgfqpoint{1.747496in}{0.938512in}}{\pgfqpoint{1.747496in}{0.916290in}}%
\pgfpathlineto{\pgfqpoint{1.747496in}{0.617535in}}%
\pgfpathquadraticcurveto{\pgfqpoint{1.747496in}{0.595313in}}{\pgfqpoint{1.769718in}{0.595313in}}%
\pgfpathclose%
\pgfusepath{stroke,fill}%
\end{pgfscope}%
\begin{pgfscope}%
\pgfsetrectcap%
\pgfsetroundjoin%
\pgfsetlinewidth{1.505625pt}%
\definecolor{currentstroke}{rgb}{1.000000,0.498039,0.054902}%
\pgfsetstrokecolor{currentstroke}%
\pgfsetdash{}{0pt}%
\pgfpathmoveto{\pgfqpoint{1.791940in}{0.855179in}}%
\pgfpathlineto{\pgfqpoint{2.014162in}{0.855179in}}%
\pgfusepath{stroke}%
\end{pgfscope}%
\begin{pgfscope}%
\pgftext[x=2.103051in,y=0.816290in,left,base]{\rmfamily\fontsize{8.000000}{9.600000}\selectfont active}%
\end{pgfscope}%
\begin{pgfscope}%
\pgfsetrectcap%
\pgfsetroundjoin%
\pgfsetlinewidth{1.505625pt}%
\definecolor{currentstroke}{rgb}{0.121569,0.466667,0.705882}%
\pgfsetstrokecolor{currentstroke}%
\pgfsetdash{}{0pt}%
\pgfpathmoveto{\pgfqpoint{1.791940in}{0.700246in}}%
\pgfpathlineto{\pgfqpoint{2.014162in}{0.700246in}}%
\pgfusepath{stroke}%
\end{pgfscope}%
\begin{pgfscope}%
\pgftext[x=2.103051in,y=0.661357in,left,base]{\rmfamily\fontsize{8.000000}{9.600000}\selectfont passive}%
\end{pgfscope}%
\end{pgfpicture}%
\makeatother%
\endgroup%

%% file: figs/e25_tradeoff.pgf
\begingroup%
\makeatletter%
\begin{pgfpicture}%
\pgfpathrectangle{\pgfpointorigin}{\pgfqpoint{2.750000in}{2.200000in}}%
\pgfusepath{use as bounding box, clip}%
\begin{pgfscope}%
\pgfsetbuttcap%
\pgfsetmiterjoin%
\definecolor{currentfill}{rgb}{1.000000,1.000000,1.000000}%
\pgfsetfillcolor{currentfill}%
\pgfsetlinewidth{0.000000pt}%
\definecolor{currentstroke}{rgb}{1.000000,1.000000,1.000000}%
\pgfsetstrokecolor{currentstroke}%
\pgfsetdash{}{0pt}%
\pgfpathmoveto{\pgfqpoint{0.000000in}{0.000000in}}%
\pgfpathlineto{\pgfqpoint{2.750000in}{0.000000in}}%
\pgfpathlineto{\pgfqpoint{2.750000in}{2.200000in}}%
\pgfpathlineto{\pgfqpoint{0.000000in}{2.200000in}}%
\pgfpathclose%
\pgfusepath{fill}%
\end{pgfscope}%
\begin{pgfscope}%
\pgfsetbuttcap%
\pgfsetmiterjoin%
\definecolor{currentfill}{rgb}{1.000000,1.000000,1.000000}%
\pgfsetfillcolor{currentfill}%
\pgfsetlinewidth{0.000000pt}%
\definecolor{currentstroke}{rgb}{0.000000,0.000000,0.000000}%
\pgfsetstrokecolor{currentstroke}%
\pgfsetstrokeopacity{0.000000}%
\pgfsetdash{}{0pt}%
\pgfpathmoveto{\pgfqpoint{0.543490in}{0.523557in}}%
\pgfpathlineto{\pgfqpoint{2.524574in}{0.523557in}}%
\pgfpathlineto{\pgfqpoint{2.524574in}{2.011738in}}%
\pgfpathlineto{\pgfqpoint{0.543490in}{2.011738in}}%
\pgfpathclose%
\pgfusepath{fill}%
\end{pgfscope}%
\begin{pgfscope}%
\pgfpathrectangle{\pgfqpoint{0.543490in}{0.523557in}}{\pgfqpoint{1.981084in}{1.488180in}} %
\pgfusepath{clip}%
\pgfsetbuttcap%
\pgfsetroundjoin%
\definecolor{currentfill}{rgb}{0.121569,0.466667,0.705882}%
\pgfsetfillcolor{currentfill}%
\pgfsetfillopacity{0.200000}%
\pgfsetlinewidth{0.000000pt}%
\definecolor{currentstroke}{rgb}{0.121569,0.466667,0.705882}%
\pgfsetstrokecolor{currentstroke}%
\pgfsetstrokeopacity{0.200000}%
\pgfsetdash{}{0pt}%
\pgfpathmoveto{\pgfqpoint{0.558150in}{0.523557in}}%
\pgfpathlineto{\pgfqpoint{0.558150in}{1.919707in}}%
\pgfpathlineto{\pgfqpoint{0.623130in}{1.883221in}}%
\pgfpathlineto{\pgfqpoint{0.912170in}{1.821199in}}%
\pgfpathlineto{\pgfqpoint{1.371385in}{1.760941in}}%
\pgfpathlineto{\pgfqpoint{1.861704in}{1.689086in}}%
\pgfpathlineto{\pgfqpoint{1.949664in}{1.587708in}}%
\pgfpathlineto{\pgfqpoint{2.243855in}{1.396657in}}%
\pgfpathlineto{\pgfqpoint{2.458802in}{1.118466in}}%
\pgfpathlineto{\pgfqpoint{2.524574in}{0.963521in}}%
\pgfpathlineto{\pgfqpoint{2.524574in}{0.523557in}}%
\pgfpathlineto{\pgfqpoint{2.524574in}{0.523557in}}%
\pgfpathlineto{\pgfqpoint{2.458802in}{0.523557in}}%
\pgfpathlineto{\pgfqpoint{2.243855in}{0.523557in}}%
\pgfpathlineto{\pgfqpoint{1.949664in}{0.523557in}}%
\pgfpathlineto{\pgfqpoint{1.861704in}{0.523557in}}%
\pgfpathlineto{\pgfqpoint{1.371385in}{0.523557in}}%
\pgfpathlineto{\pgfqpoint{0.912170in}{0.523557in}}%
\pgfpathlineto{\pgfqpoint{0.623130in}{0.523557in}}%
\pgfpathlineto{\pgfqpoint{0.558150in}{0.523557in}}%
\pgfpathclose%
\pgfusepath{fill}%
\end{pgfscope}%
\begin{pgfscope}%
\pgfpathrectangle{\pgfqpoint{0.543490in}{0.523557in}}{\pgfqpoint{1.981084in}{1.488180in}} %
\pgfusepath{clip}%
\pgfsetbuttcap%
\pgfsetroundjoin%
\pgfsetlinewidth{0.000000pt}%
\definecolor{currentstroke}{rgb}{1.000000,0.498039,0.054902}%
\pgfsetstrokecolor{currentstroke}%
\pgfsetstrokeopacity{0.200000}%
\pgfsetdash{}{0pt}%
\pgfpathmoveto{\pgfqpoint{0.558546in}{0.523557in}}%
\pgfpathlineto{\pgfqpoint{0.558546in}{1.916227in}}%
\pgfpathlineto{\pgfqpoint{0.635611in}{1.880077in}}%
\pgfpathlineto{\pgfqpoint{0.799446in}{1.836880in}}%
\pgfpathlineto{\pgfqpoint{1.127514in}{1.808530in}}%
\pgfpathlineto{\pgfqpoint{1.722632in}{1.759259in}}%
\pgfpathlineto{\pgfqpoint{2.069717in}{1.720548in}}%
\pgfpathlineto{\pgfqpoint{2.331221in}{1.682095in}}%
\pgfpathlineto{\pgfqpoint{2.495453in}{1.665987in}}%
\pgfpathlineto{\pgfqpoint{2.522395in}{1.642768in}}%
\pgfpathlineto{\pgfqpoint{2.524574in}{1.618020in}}%
\pgfpathlineto{\pgfqpoint{2.524574in}{0.523557in}}%
\pgfpathlineto{\pgfqpoint{2.524574in}{0.523557in}}%
\pgfpathlineto{\pgfqpoint{2.522395in}{0.523557in}}%
\pgfpathlineto{\pgfqpoint{2.495453in}{0.523557in}}%
\pgfpathlineto{\pgfqpoint{2.331221in}{0.523557in}}%
\pgfpathlineto{\pgfqpoint{2.069717in}{0.523557in}}%
\pgfpathlineto{\pgfqpoint{1.722632in}{0.523557in}}%
\pgfpathlineto{\pgfqpoint{1.127514in}{0.523557in}}%
\pgfpathlineto{\pgfqpoint{0.799446in}{0.523557in}}%
\pgfpathlineto{\pgfqpoint{0.635611in}{0.523557in}}%
\pgfpathlineto{\pgfqpoint{0.558546in}{0.523557in}}%
\pgfpathclose%
\pgfusepath{}%
\end{pgfscope}%
\begin{pgfscope}%
\pgfsetbuttcap%
\pgfsetroundjoin%
\pgfsetlinewidth{0.000000pt}%
\definecolor{currentstroke}{rgb}{1.000000,0.498039,0.054902}%
\pgfsetstrokecolor{currentstroke}%
\pgfsetstrokeopacity{0.200000}%
\pgfsetdash{}{0pt}%
\pgfpathrectangle{\pgfqpoint{0.543490in}{0.523557in}}{\pgfqpoint{1.981084in}{1.488180in}} %
\pgfusepath{clip}%
\pgfpathmoveto{\pgfqpoint{0.558546in}{0.523557in}}%
\pgfpathlineto{\pgfqpoint{0.558546in}{1.916227in}}%
\pgfpathlineto{\pgfqpoint{0.635611in}{1.880077in}}%
\pgfpathlineto{\pgfqpoint{0.799446in}{1.836880in}}%
\pgfpathlineto{\pgfqpoint{1.127514in}{1.808530in}}%
\pgfpathlineto{\pgfqpoint{1.722632in}{1.759259in}}%
\pgfpathlineto{\pgfqpoint{2.069717in}{1.720548in}}%
\pgfpathlineto{\pgfqpoint{2.331221in}{1.682095in}}%
\pgfpathlineto{\pgfqpoint{2.495453in}{1.665987in}}%
\pgfpathlineto{\pgfqpoint{2.522395in}{1.642768in}}%
\pgfpathlineto{\pgfqpoint{2.524574in}{1.618020in}}%
\pgfpathlineto{\pgfqpoint{2.524574in}{0.523557in}}%
\pgfpathlineto{\pgfqpoint{2.524574in}{0.523557in}}%
\pgfpathlineto{\pgfqpoint{2.522395in}{0.523557in}}%
\pgfpathlineto{\pgfqpoint{2.495453in}{0.523557in}}%
\pgfpathlineto{\pgfqpoint{2.331221in}{0.523557in}}%
\pgfpathlineto{\pgfqpoint{2.069717in}{0.523557in}}%
\pgfpathlineto{\pgfqpoint{1.722632in}{0.523557in}}%
\pgfpathlineto{\pgfqpoint{1.127514in}{0.523557in}}%
\pgfpathlineto{\pgfqpoint{0.799446in}{0.523557in}}%
\pgfpathlineto{\pgfqpoint{0.635611in}{0.523557in}}%
\pgfpathlineto{\pgfqpoint{0.558546in}{0.523557in}}%
\pgfpathclose%
\pgfusepath{clip}%
\pgfsys@defobject{currentpattern}{\pgfqpoint{0in}{0in}}{\pgfqpoint{1in}{1in}}{%
\begin{pgfscope}%
\pgfpathrectangle{\pgfqpoint{0in}{0in}}{\pgfqpoint{1in}{1in}}%
\pgfusepath{clip}%
\pgfpathmoveto{\pgfqpoint{-0.500000in}{0.500000in}}%
\pgfpathlineto{\pgfqpoint{0.500000in}{1.500000in}}%
\pgfpathmoveto{\pgfqpoint{-0.416667in}{0.416667in}}%
\pgfpathlineto{\pgfqpoint{0.583333in}{1.416667in}}%
\pgfpathmoveto{\pgfqpoint{-0.333333in}{0.333333in}}%
\pgfpathlineto{\pgfqpoint{0.666667in}{1.333333in}}%
\pgfpathmoveto{\pgfqpoint{-0.250000in}{0.250000in}}%
\pgfpathlineto{\pgfqpoint{0.750000in}{1.250000in}}%
\pgfpathmoveto{\pgfqpoint{-0.166667in}{0.166667in}}%
\pgfpathlineto{\pgfqpoint{0.833333in}{1.166667in}}%
\pgfpathmoveto{\pgfqpoint{-0.083333in}{0.083333in}}%
\pgfpathlineto{\pgfqpoint{0.916667in}{1.083333in}}%
\pgfpathmoveto{\pgfqpoint{0.000000in}{0.000000in}}%
\pgfpathlineto{\pgfqpoint{1.000000in}{1.000000in}}%
\pgfpathmoveto{\pgfqpoint{0.083333in}{-0.083333in}}%
\pgfpathlineto{\pgfqpoint{1.083333in}{0.916667in}}%
\pgfpathmoveto{\pgfqpoint{0.166667in}{-0.166667in}}%
\pgfpathlineto{\pgfqpoint{1.166667in}{0.833333in}}%
\pgfpathmoveto{\pgfqpoint{0.250000in}{-0.250000in}}%
\pgfpathlineto{\pgfqpoint{1.250000in}{0.750000in}}%
\pgfpathmoveto{\pgfqpoint{0.333333in}{-0.333333in}}%
\pgfpathlineto{\pgfqpoint{1.333333in}{0.666667in}}%
\pgfpathmoveto{\pgfqpoint{0.416667in}{-0.416667in}}%
\pgfpathlineto{\pgfqpoint{1.416667in}{0.583333in}}%
\pgfpathmoveto{\pgfqpoint{0.500000in}{-0.500000in}}%
\pgfpathlineto{\pgfqpoint{1.500000in}{0.500000in}}%
\pgfusepath{stroke}%
\end{pgfscope}%
}%
\pgfsys@transformshift{0.558546in}{0.523557in}%
\pgfsys@useobject{currentpattern}{}%
\pgfsys@transformshift{1in}{0in}%
\pgfsys@useobject{currentpattern}{}%
\pgfsys@transformshift{1in}{0in}%
\pgfsys@transformshift{-2in}{0in}%
\pgfsys@transformshift{0in}{1in}%
\pgfsys@useobject{currentpattern}{}%
\pgfsys@transformshift{1in}{0in}%
\pgfsys@useobject{currentpattern}{}%
\pgfsys@transformshift{1in}{0in}%
\pgfsys@transformshift{-2in}{0in}%
\pgfsys@transformshift{0in}{1in}%
\end{pgfscope}%
\begin{pgfscope}%
\pgfsetbuttcap%
\pgfsetroundjoin%
\definecolor{currentfill}{rgb}{0.000000,0.000000,0.000000}%
\pgfsetfillcolor{currentfill}%
\pgfsetlinewidth{0.803000pt}%
\definecolor{currentstroke}{rgb}{0.000000,0.000000,0.000000}%
\pgfsetstrokecolor{currentstroke}%
\pgfsetdash{}{0pt}%
\pgfsys@defobject{currentmarker}{\pgfqpoint{0.000000in}{-0.048611in}}{\pgfqpoint{0.000000in}{0.000000in}}{%
\pgfpathmoveto{\pgfqpoint{0.000000in}{0.000000in}}%
\pgfpathlineto{\pgfqpoint{0.000000in}{-0.048611in}}%
\pgfusepath{stroke,fill}%
}%
\begin{pgfscope}%
\pgfsys@transformshift{0.543490in}{0.523557in}%
\pgfsys@useobject{currentmarker}{}%
\end{pgfscope}%
\end{pgfscope}%
\begin{pgfscope}%
\pgftext[x=0.543490in,y=0.426335in,,top]{\rmfamily\fontsize{8.000000}{9.600000}\selectfont \(\displaystyle 0.0\)}%
\end{pgfscope}%
\begin{pgfscope}%
\pgfsetbuttcap%
\pgfsetroundjoin%
\definecolor{currentfill}{rgb}{0.000000,0.000000,0.000000}%
\pgfsetfillcolor{currentfill}%
\pgfsetlinewidth{0.803000pt}%
\definecolor{currentstroke}{rgb}{0.000000,0.000000,0.000000}%
\pgfsetstrokecolor{currentstroke}%
\pgfsetdash{}{0pt}%
\pgfsys@defobject{currentmarker}{\pgfqpoint{0.000000in}{-0.048611in}}{\pgfqpoint{0.000000in}{0.000000in}}{%
\pgfpathmoveto{\pgfqpoint{0.000000in}{0.000000in}}%
\pgfpathlineto{\pgfqpoint{0.000000in}{-0.048611in}}%
\pgfusepath{stroke,fill}%
}%
\begin{pgfscope}%
\pgfsys@transformshift{0.939707in}{0.523557in}%
\pgfsys@useobject{currentmarker}{}%
\end{pgfscope}%
\end{pgfscope}%
\begin{pgfscope}%
\pgftext[x=0.939707in,y=0.426335in,,top]{\rmfamily\fontsize{8.000000}{9.600000}\selectfont \(\displaystyle 0.2\)}%
\end{pgfscope}%
\begin{pgfscope}%
\pgfsetbuttcap%
\pgfsetroundjoin%
\definecolor{currentfill}{rgb}{0.000000,0.000000,0.000000}%
\pgfsetfillcolor{currentfill}%
\pgfsetlinewidth{0.803000pt}%
\definecolor{currentstroke}{rgb}{0.000000,0.000000,0.000000}%
\pgfsetstrokecolor{currentstroke}%
\pgfsetdash{}{0pt}%
\pgfsys@defobject{currentmarker}{\pgfqpoint{0.000000in}{-0.048611in}}{\pgfqpoint{0.000000in}{0.000000in}}{%
\pgfpathmoveto{\pgfqpoint{0.000000in}{0.000000in}}%
\pgfpathlineto{\pgfqpoint{0.000000in}{-0.048611in}}%
\pgfusepath{stroke,fill}%
}%
\begin{pgfscope}%
\pgfsys@transformshift{1.335924in}{0.523557in}%
\pgfsys@useobject{currentmarker}{}%
\end{pgfscope}%
\end{pgfscope}%
\begin{pgfscope}%
\pgftext[x=1.335924in,y=0.426335in,,top]{\rmfamily\fontsize{8.000000}{9.600000}\selectfont \(\displaystyle 0.4\)}%
\end{pgfscope}%
\begin{pgfscope}%
\pgfsetbuttcap%
\pgfsetroundjoin%
\definecolor{currentfill}{rgb}{0.000000,0.000000,0.000000}%
\pgfsetfillcolor{currentfill}%
\pgfsetlinewidth{0.803000pt}%
\definecolor{currentstroke}{rgb}{0.000000,0.000000,0.000000}%
\pgfsetstrokecolor{currentstroke}%
\pgfsetdash{}{0pt}%
\pgfsys@defobject{currentmarker}{\pgfqpoint{0.000000in}{-0.048611in}}{\pgfqpoint{0.000000in}{0.000000in}}{%
\pgfpathmoveto{\pgfqpoint{0.000000in}{0.000000in}}%
\pgfpathlineto{\pgfqpoint{0.000000in}{-0.048611in}}%
\pgfusepath{stroke,fill}%
}%
\begin{pgfscope}%
\pgfsys@transformshift{1.732141in}{0.523557in}%
\pgfsys@useobject{currentmarker}{}%
\end{pgfscope}%
\end{pgfscope}%
\begin{pgfscope}%
\pgftext[x=1.732141in,y=0.426335in,,top]{\rmfamily\fontsize{8.000000}{9.600000}\selectfont \(\displaystyle 0.6\)}%
\end{pgfscope}%
\begin{pgfscope}%
\pgfsetbuttcap%
\pgfsetroundjoin%
\definecolor{currentfill}{rgb}{0.000000,0.000000,0.000000}%
\pgfsetfillcolor{currentfill}%
\pgfsetlinewidth{0.803000pt}%
\definecolor{currentstroke}{rgb}{0.000000,0.000000,0.000000}%
\pgfsetstrokecolor{currentstroke}%
\pgfsetdash{}{0pt}%
\pgfsys@defobject{currentmarker}{\pgfqpoint{0.000000in}{-0.048611in}}{\pgfqpoint{0.000000in}{0.000000in}}{%
\pgfpathmoveto{\pgfqpoint{0.000000in}{0.000000in}}%
\pgfpathlineto{\pgfqpoint{0.000000in}{-0.048611in}}%
\pgfusepath{stroke,fill}%
}%
\begin{pgfscope}%
\pgfsys@transformshift{2.128358in}{0.523557in}%
\pgfsys@useobject{currentmarker}{}%
\end{pgfscope}%
\end{pgfscope}%
\begin{pgfscope}%
\pgftext[x=2.128358in,y=0.426335in,,top]{\rmfamily\fontsize{8.000000}{9.600000}\selectfont \(\displaystyle 0.8\)}%
\end{pgfscope}%
\begin{pgfscope}%
\pgfsetbuttcap%
\pgfsetroundjoin%
\definecolor{currentfill}{rgb}{0.000000,0.000000,0.000000}%
\pgfsetfillcolor{currentfill}%
\pgfsetlinewidth{0.803000pt}%
\definecolor{currentstroke}{rgb}{0.000000,0.000000,0.000000}%
\pgfsetstrokecolor{currentstroke}%
\pgfsetdash{}{0pt}%
\pgfsys@defobject{currentmarker}{\pgfqpoint{0.000000in}{-0.048611in}}{\pgfqpoint{0.000000in}{0.000000in}}{%
\pgfpathmoveto{\pgfqpoint{0.000000in}{0.000000in}}%
\pgfpathlineto{\pgfqpoint{0.000000in}{-0.048611in}}%
\pgfusepath{stroke,fill}%
}%
\begin{pgfscope}%
\pgfsys@transformshift{2.524574in}{0.523557in}%
\pgfsys@useobject{currentmarker}{}%
\end{pgfscope}%
\end{pgfscope}%
\begin{pgfscope}%
\pgftext[x=2.524574in,y=0.426335in,,top]{\rmfamily\fontsize{8.000000}{9.600000}\selectfont \(\displaystyle 1.0\)}%
\end{pgfscope}%
\begin{pgfscope}%
\pgftext[x=1.534032in,y=0.272655in,,top]{\rmfamily\fontsize{10.000000}{12.000000}\selectfont validity fraction of samples}%
\end{pgfscope}%
\begin{pgfscope}%
\pgfsetbuttcap%
\pgfsetroundjoin%
\definecolor{currentfill}{rgb}{0.000000,0.000000,0.000000}%
\pgfsetfillcolor{currentfill}%
\pgfsetlinewidth{0.803000pt}%
\definecolor{currentstroke}{rgb}{0.000000,0.000000,0.000000}%
\pgfsetstrokecolor{currentstroke}%
\pgfsetdash{}{0pt}%
\pgfsys@defobject{currentmarker}{\pgfqpoint{-0.048611in}{0.000000in}}{\pgfqpoint{0.000000in}{0.000000in}}{%
\pgfpathmoveto{\pgfqpoint{0.000000in}{0.000000in}}%
\pgfpathlineto{\pgfqpoint{-0.048611in}{0.000000in}}%
\pgfusepath{stroke,fill}%
}%
\begin{pgfscope}%
\pgfsys@transformshift{0.543490in}{0.523557in}%
\pgfsys@useobject{currentmarker}{}%
\end{pgfscope}%
\end{pgfscope}%
\begin{pgfscope}%
\pgftext[x=0.387239in,y=0.485295in,left,base]{\rmfamily\fontsize{8.000000}{9.600000}\selectfont \(\displaystyle 0\)}%
\end{pgfscope}%
\begin{pgfscope}%
\pgfsetbuttcap%
\pgfsetroundjoin%
\definecolor{currentfill}{rgb}{0.000000,0.000000,0.000000}%
\pgfsetfillcolor{currentfill}%
\pgfsetlinewidth{0.803000pt}%
\definecolor{currentstroke}{rgb}{0.000000,0.000000,0.000000}%
\pgfsetstrokecolor{currentstroke}%
\pgfsetdash{}{0pt}%
\pgfsys@defobject{currentmarker}{\pgfqpoint{-0.048611in}{0.000000in}}{\pgfqpoint{0.000000in}{0.000000in}}{%
\pgfpathmoveto{\pgfqpoint{0.000000in}{0.000000in}}%
\pgfpathlineto{\pgfqpoint{-0.048611in}{0.000000in}}%
\pgfusepath{stroke,fill}%
}%
\begin{pgfscope}%
\pgfsys@transformshift{0.543490in}{0.895602in}%
\pgfsys@useobject{currentmarker}{}%
\end{pgfscope}%
\end{pgfscope}%
\begin{pgfscope}%
\pgftext[x=0.328211in,y=0.857340in,left,base]{\rmfamily\fontsize{8.000000}{9.600000}\selectfont \(\displaystyle 20\)}%
\end{pgfscope}%
\begin{pgfscope}%
\pgfsetbuttcap%
\pgfsetroundjoin%
\definecolor{currentfill}{rgb}{0.000000,0.000000,0.000000}%
\pgfsetfillcolor{currentfill}%
\pgfsetlinewidth{0.803000pt}%
\definecolor{currentstroke}{rgb}{0.000000,0.000000,0.000000}%
\pgfsetstrokecolor{currentstroke}%
\pgfsetdash{}{0pt}%
\pgfsys@defobject{currentmarker}{\pgfqpoint{-0.048611in}{0.000000in}}{\pgfqpoint{0.000000in}{0.000000in}}{%
\pgfpathmoveto{\pgfqpoint{0.000000in}{0.000000in}}%
\pgfpathlineto{\pgfqpoint{-0.048611in}{0.000000in}}%
\pgfusepath{stroke,fill}%
}%
\begin{pgfscope}%
\pgfsys@transformshift{0.543490in}{1.267647in}%
\pgfsys@useobject{currentmarker}{}%
\end{pgfscope}%
\end{pgfscope}%
\begin{pgfscope}%
\pgftext[x=0.328211in,y=1.229385in,left,base]{\rmfamily\fontsize{8.000000}{9.600000}\selectfont \(\displaystyle 40\)}%
\end{pgfscope}%
\begin{pgfscope}%
\pgfsetbuttcap%
\pgfsetroundjoin%
\definecolor{currentfill}{rgb}{0.000000,0.000000,0.000000}%
\pgfsetfillcolor{currentfill}%
\pgfsetlinewidth{0.803000pt}%
\definecolor{currentstroke}{rgb}{0.000000,0.000000,0.000000}%
\pgfsetstrokecolor{currentstroke}%
\pgfsetdash{}{0pt}%
\pgfsys@defobject{currentmarker}{\pgfqpoint{-0.048611in}{0.000000in}}{\pgfqpoint{0.000000in}{0.000000in}}{%
\pgfpathmoveto{\pgfqpoint{0.000000in}{0.000000in}}%
\pgfpathlineto{\pgfqpoint{-0.048611in}{0.000000in}}%
\pgfusepath{stroke,fill}%
}%
\begin{pgfscope}%
\pgfsys@transformshift{0.543490in}{1.639693in}%
\pgfsys@useobject{currentmarker}{}%
\end{pgfscope}%
\end{pgfscope}%
\begin{pgfscope}%
\pgftext[x=0.328211in,y=1.601430in,left,base]{\rmfamily\fontsize{8.000000}{9.600000}\selectfont \(\displaystyle 60\)}%
\end{pgfscope}%
\begin{pgfscope}%
\pgfsetbuttcap%
\pgfsetroundjoin%
\definecolor{currentfill}{rgb}{0.000000,0.000000,0.000000}%
\pgfsetfillcolor{currentfill}%
\pgfsetlinewidth{0.803000pt}%
\definecolor{currentstroke}{rgb}{0.000000,0.000000,0.000000}%
\pgfsetstrokecolor{currentstroke}%
\pgfsetdash{}{0pt}%
\pgfsys@defobject{currentmarker}{\pgfqpoint{-0.048611in}{0.000000in}}{\pgfqpoint{0.000000in}{0.000000in}}{%
\pgfpathmoveto{\pgfqpoint{0.000000in}{0.000000in}}%
\pgfpathlineto{\pgfqpoint{-0.048611in}{0.000000in}}%
\pgfusepath{stroke,fill}%
}%
\begin{pgfscope}%
\pgfsys@transformshift{0.543490in}{2.011738in}%
\pgfsys@useobject{currentmarker}{}%
\end{pgfscope}%
\end{pgfscope}%
\begin{pgfscope}%
\pgftext[x=0.328211in,y=1.973475in,left,base]{\rmfamily\fontsize{8.000000}{9.600000}\selectfont \(\displaystyle 80\)}%
\end{pgfscope}%
\begin{pgfscope}%
\pgftext[x=0.272655in,y=1.267647in,,bottom,rotate=90.000000]{\rmfamily\fontsize{10.000000}{12.000000}\selectfont entropy}%
\end{pgfscope}%
\begin{pgfscope}%
\pgfpathrectangle{\pgfqpoint{0.543490in}{0.523557in}}{\pgfqpoint{1.981084in}{1.488180in}} %
\pgfusepath{clip}%
\pgfsetrectcap%
\pgfsetroundjoin%
\pgfsetlinewidth{1.505625pt}%
\definecolor{currentstroke}{rgb}{0.121569,0.466667,0.705882}%
\pgfsetstrokecolor{currentstroke}%
\pgfsetdash{}{0pt}%
\pgfpathmoveto{\pgfqpoint{0.558150in}{1.919707in}}%
\pgfpathlineto{\pgfqpoint{0.623130in}{1.883221in}}%
\pgfpathlineto{\pgfqpoint{0.912170in}{1.821199in}}%
\pgfpathlineto{\pgfqpoint{1.371385in}{1.760941in}}%
\pgfpathlineto{\pgfqpoint{1.861704in}{1.689086in}}%
\pgfpathlineto{\pgfqpoint{1.949664in}{1.587708in}}%
\pgfpathlineto{\pgfqpoint{2.243855in}{1.396657in}}%
\pgfpathlineto{\pgfqpoint{2.458802in}{1.118466in}}%
\pgfpathlineto{\pgfqpoint{2.524574in}{0.963521in}}%
\pgfusepath{stroke}%
\end{pgfscope}%
\begin{pgfscope}%
\pgfpathrectangle{\pgfqpoint{0.543490in}{0.523557in}}{\pgfqpoint{1.981084in}{1.488180in}} %
\pgfusepath{clip}%
\pgfsetrectcap%
\pgfsetroundjoin%
\pgfsetlinewidth{1.505625pt}%
\definecolor{currentstroke}{rgb}{1.000000,0.498039,0.054902}%
\pgfsetstrokecolor{currentstroke}%
\pgfsetdash{}{0pt}%
\pgfpathmoveto{\pgfqpoint{0.558546in}{1.916227in}}%
\pgfpathlineto{\pgfqpoint{0.635611in}{1.880077in}}%
\pgfpathlineto{\pgfqpoint{0.799446in}{1.836880in}}%
\pgfpathlineto{\pgfqpoint{1.127514in}{1.808530in}}%
\pgfpathlineto{\pgfqpoint{1.722632in}{1.759259in}}%
\pgfpathlineto{\pgfqpoint{2.069717in}{1.720548in}}%
\pgfpathlineto{\pgfqpoint{2.331221in}{1.682095in}}%
\pgfpathlineto{\pgfqpoint{2.495453in}{1.665987in}}%
\pgfpathlineto{\pgfqpoint{2.522395in}{1.642768in}}%
\pgfpathlineto{\pgfqpoint{2.524574in}{1.618020in}}%
\pgfusepath{stroke}%
\end{pgfscope}%
\begin{pgfscope}%
\pgfsetrectcap%
\pgfsetmiterjoin%
\pgfsetlinewidth{0.803000pt}%
\definecolor{currentstroke}{rgb}{0.000000,0.000000,0.000000}%
\pgfsetstrokecolor{currentstroke}%
\pgfsetdash{}{0pt}%
\pgfpathmoveto{\pgfqpoint{0.543490in}{0.523557in}}%
\pgfpathlineto{\pgfqpoint{0.543490in}{2.011738in}}%
\pgfusepath{stroke}%
\end{pgfscope}%
\begin{pgfscope}%
\pgfsetrectcap%
\pgfsetmiterjoin%
\pgfsetlinewidth{0.803000pt}%
\definecolor{currentstroke}{rgb}{0.000000,0.000000,0.000000}%
\pgfsetstrokecolor{currentstroke}%
\pgfsetdash{}{0pt}%
\pgfpathmoveto{\pgfqpoint{2.524574in}{0.523557in}}%
\pgfpathlineto{\pgfqpoint{2.524574in}{2.011738in}}%
\pgfusepath{stroke}%
\end{pgfscope}%
\begin{pgfscope}%
\pgfsetrectcap%
\pgfsetmiterjoin%
\pgfsetlinewidth{0.803000pt}%
\definecolor{currentstroke}{rgb}{0.000000,0.000000,0.000000}%
\pgfsetstrokecolor{currentstroke}%
\pgfsetdash{}{0pt}%
\pgfpathmoveto{\pgfqpoint{0.543490in}{0.523557in}}%
\pgfpathlineto{\pgfqpoint{2.524574in}{0.523557in}}%
\pgfusepath{stroke}%
\end{pgfscope}%
\begin{pgfscope}%
\pgfsetrectcap%
\pgfsetmiterjoin%
\pgfsetlinewidth{0.803000pt}%
\definecolor{currentstroke}{rgb}{0.000000,0.000000,0.000000}%
\pgfsetstrokecolor{currentstroke}%
\pgfsetdash{}{0pt}%
\pgfpathmoveto{\pgfqpoint{0.543490in}{2.011738in}}%
\pgfpathlineto{\pgfqpoint{2.524574in}{2.011738in}}%
\pgfusepath{stroke}%
\end{pgfscope}%
\begin{pgfscope}%
\pgfsetbuttcap%
\pgfsetmiterjoin%
\definecolor{currentfill}{rgb}{1.000000,1.000000,1.000000}%
\pgfsetfillcolor{currentfill}%
\pgfsetlinewidth{1.003750pt}%
\definecolor{currentstroke}{rgb}{0.800000,0.800000,0.800000}%
\pgfsetstrokecolor{currentstroke}%
\pgfsetdash{}{0pt}%
\pgfpathmoveto{\pgfqpoint{1.729293in}{0.579113in}}%
\pgfpathlineto{\pgfqpoint{2.446797in}{0.579113in}}%
\pgfpathquadraticcurveto{\pgfqpoint{2.469019in}{0.579113in}}{\pgfqpoint{2.469019in}{0.601335in}}%
\pgfpathlineto{\pgfqpoint{2.469019in}{0.900090in}}%
\pgfpathquadraticcurveto{\pgfqpoint{2.469019in}{0.922312in}}{\pgfqpoint{2.446797in}{0.922312in}}%
\pgfpathlineto{\pgfqpoint{1.729293in}{0.922312in}}%
\pgfpathquadraticcurveto{\pgfqpoint{1.707070in}{0.922312in}}{\pgfqpoint{1.707070in}{0.900090in}}%
\pgfpathlineto{\pgfqpoint{1.707070in}{0.601335in}}%
\pgfpathquadraticcurveto{\pgfqpoint{1.707070in}{0.579113in}}{\pgfqpoint{1.729293in}{0.579113in}}%
\pgfpathclose%
\pgfusepath{stroke,fill}%
\end{pgfscope}%
\begin{pgfscope}%
\pgfsetrectcap%
\pgfsetroundjoin%
\pgfsetlinewidth{1.505625pt}%
\definecolor{currentstroke}{rgb}{1.000000,0.498039,0.054902}%
\pgfsetstrokecolor{currentstroke}%
\pgfsetdash{}{0pt}%
\pgfpathmoveto{\pgfqpoint{1.751515in}{0.838979in}}%
\pgfpathlineto{\pgfqpoint{1.973737in}{0.838979in}}%
\pgfusepath{stroke}%
\end{pgfscope}%
\begin{pgfscope}%
\pgftext[x=2.062626in,y=0.800090in,left,base]{\rmfamily\fontsize{8.000000}{9.600000}\selectfont active}%
\end{pgfscope}%
\begin{pgfscope}%
\pgfsetrectcap%
\pgfsetroundjoin%
\pgfsetlinewidth{1.505625pt}%
\definecolor{currentstroke}{rgb}{0.121569,0.466667,0.705882}%
\pgfsetstrokecolor{currentstroke}%
\pgfsetdash{}{0pt}%
\pgfpathmoveto{\pgfqpoint{1.751515in}{0.684046in}}%
\pgfpathlineto{\pgfqpoint{1.973737in}{0.684046in}}%
\pgfusepath{stroke}%
\end{pgfscope}%
\begin{pgfscope}%
\pgftext[x=2.062626in,y=0.645157in,left,base]{\rmfamily\fontsize{8.000000}{9.600000}\selectfont passive}%
\end{pgfscope}%
\end{pgfpicture}%
\makeatother%
\endgroup%

%% file: sections/molecules.tex
\subsection{SMILES molecules}
\begin{figure}[t]
\centering
\includegraphics[width=\textwidth]{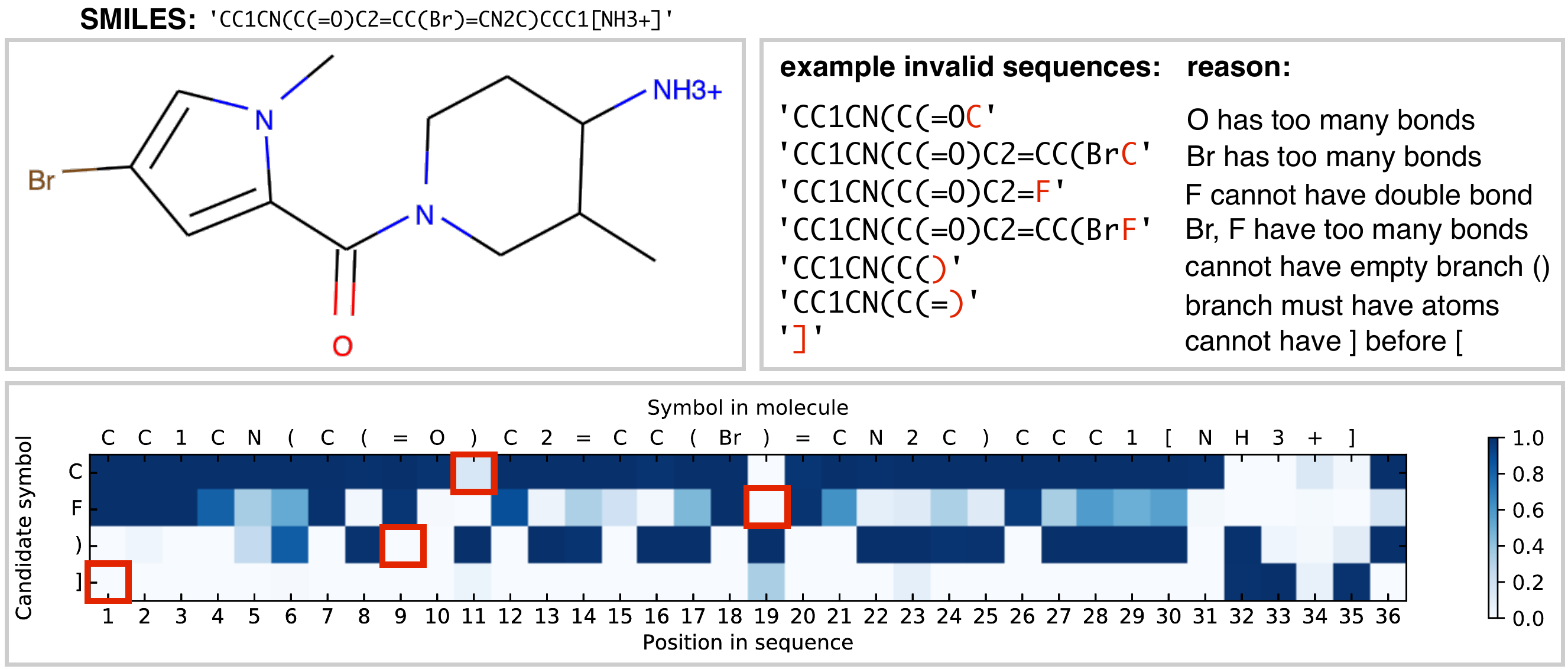}
\caption{Predictions $\out$ of the agent at each step $t$ for the valid test
  molecule shown in the top left figure, for a subset of possible actions
  (selecting as next character \texttt{C}, \texttt{F}, \texttt{)}, or
  \texttt{]}). Each column shows which actions the trained agent believes are
  valid at each $t$, given the characters $x_{<t}$ preceding it. We see that the
  validity model has learned basic valence constraints: for example the oxygen
  atom \texttt{O} at position 10 can form at most 2 bonds, and since it is
  preceded by a double bond, the model knows that neither carbon \texttt{C} nor
  fluorine \texttt{F} can immediately follow it at position 11; we see the same
  after the bromine \texttt{Br} at position 18, which can only form a single
  bond. The model also correctly identifies that closing branch symbols
  \texttt{)} cannot immediately follow opening branches (after positions 6, 8,
  and 17), as well as that closing brackets \texttt{]} cannot occur until an
  open bracket has been followed by at least one atom (at positions 32--35). The
  full output heatmap for this example molecule is shown in
  Figure~\ref{fig:full-heatmap} in the appendix. }
\label{fig:example-mol}
\end{figure}

SMILES strings \citep{weininger1970smiles} are one of the most common
representations for molecules, consisting of an ordering of atoms and
bonds. It is attractive for many applications because it maps the graphical
representation of a molecule to a sequential representation, capturing not just
its chemical composition but also structure. This structural information
is captured by intricate dependencies in SMILES strings based on chemical
properties of individual atoms and valid atom connectivities. For
instance, the atom Bromine can only bond with a single other atom, meaning that it may only occur at the beginning or end of a SMILES string, or within a
so-called `branch', denoted by a bracketed expression \texttt{(Br)}. We
illustrate some of these rules, including a Bromine branch, in
figure~\ref{fig:example-mol}, with a graphical representation of a molecule alongside its corresponding SMILES string. There, we also show examples of how a string
may fail to form a valid SMILES molecule representation. The full SMILES
alphabet is presented in table~\ref{tab:smiles-alphabet}.
\begin{table}[hb]
  \centering
  \caption{SMILES alphabet}
  \label{tab:smiles-alphabet}
  \begin{tabular}{c c c c}
    \toprule
  atoms/chirality & bonds/ringbonds & charges & branches/brackets \\
  \midrule
\texttt{B} \texttt{C} \texttt{N} \texttt{O} \texttt{S} \texttt{P} \texttt{F} \texttt{I} \texttt{H} \texttt{Cl} \texttt{Br} \texttt{@}  & \texttt{=\#/\textbackslash12345678} & \texttt{-+} & \texttt{()[]} \\
  \bottomrule
\end{tabular}
\end{table}

The intricacy of SMILES strings makes them a suitable testing ground for our
method. There are two technical distinctions to make between this experimental
setup and the previously considered Python 3 mathematical expressions. As there
exist databases of SMILES strings, we leverage those by using the data
augmentation technique described in section \ref{sec:aug}. The main data source
considered is the ZINC data set \cite{zinc}, as used in \cite{kusner17}. We also
use the USPTO 15k reaction products data \citep{uspto} and a set of molecule
solubility information \citep{solubility} as withheld test data.
Secondly, whereas we used fixed length Python 3 expressions in order to obtain
coverage bounds, molecules are inherently of variable length. We deal with this by
padding all molecules to fixed length.

\paragraph{Validating grammar model accuracy} As a first test of the suitability
of our proposed validity model, we train it on augmented ZINC data and examine
the accuracy of its predictions on a withheld test partition of that same data
set as well as the two unseen molecule data sets. Accuracy is the ability of the
model to accurately recognise which perturbations make a certain SMILES string
invalid, and which leave it valid -- effectively how well the model has captured
the grammar of SMILES strings in the vicinity of the data manifold. Recalling
that a sequence is invalid if $\tilde{v}(x_{1:t}) = 0$ at any $t \leq T$, we
consider the model prediction for molecule $x_{1:T}$ to be
$\prod_{t=1}^T \mathbb{I}\left[\out \geq 0.5\right]$, and compare this to its
true label as given by rdkit, a chemical informatics software. The results are
encouraging, with the model achieving 0.998 accuracy on perturbed ZINC (test)
and 1.000 accuracy on both perturbed USPTO and perturbed Solubility withheld
data. Perturbation rate was selected such that approximately half of the
perturbed strings are valid.

\paragraph{Integrating with Character VAE} To demonstrate the models capability
of improving preexisting generative models for discrete structures, we show how
it can be used to improve the results of previous work, a character variational
autoencoder (CVAE) applied to SMILES strings
\citep{gomez-bombarelli_automatic_2016, kingma2013auto}. Therein, an encoder
maps points in $\inspace_+$ to a continuous latent representation $\mathcal{Z}$
and a paired decoder maps points in $\mathcal{Z}$ back to $\inspace_+$. A
reconstruction based loss is minimised such that training points mapped to the
latent space decode back into the same SMILES strings. The fraction of test
points that do is termed reconstruction accuracy. The loss also features a term
that encourages the posterior over $\mathcal{Z}$ to be close to some prior,
typically a normal distribution. A key metric for the performance of variational
autoencoder models for discrete structures is the fraction of points sampled
from the prior over $\mathcal{Z}$ that decode into valid molecules. If many
points do not correspond to valid molecules, any sort of predictive modeling on
that space will likely also mostly output invalid SMILES strings.

The decoder functions by outputting a set of weights $f(x_t | z)$ for each
character $x_t$ in the reconstructed sequence conditioned on a latent point
$z\in\mathcal{Z}$; the sequence is recovered by sampling from a multinomial
according to these weights. To integrate our validity model into this framework,
we take the decoder output for each step $t$ and mask out choices that we
predict cannot give valid sequence continuations. We thus sample characters with
weights given by $f(x_t | z) \cdot \mathbb{I}\left[\out \geq 0.5\right]$.

\paragraph{Autoencoding benchmarks} Table \ref{tbl:vae} contains a comparison of
our work to a plain CVAE and to the Grammar VAE approach. We use a Kekul\'{e}
format of the ZINC data in our experiments, a specific representation of
aromatic bonds that our model handled particularly well. Note that the results
we quote for Grammar VAE are taken directly from \cite{kusner17} and on
non-Kekul\'{e} format data. The CVAE model is trained for 100 epochs, as
per previous work -- further training improves reconstruction accuracy.

We note that the binary nature of the proposed grammar model means that it does
not affect the reconstruction accuracy. In fact, some modest gains are present.
The addition of our grammar model to the character VAE significantly improves
its ability to decode discrete structures, as seen by the order of magnitude
increase in latent sample validity. The action of our model is completely
post-hoc and thus can be applied to any pre-trained character-based VAE model
where elements of the latent space correspond to a structured discrete sequence.
\begin{table}[h]
  \centering
  \begin{tabular}[h]{c c c}
    \toprule
    Model & reconstruction accuracy & sample validity \\
    \midrule
    CVAE + Validity Model & 50.2\% & 22.3\% \\
    Grammar VAE & 53.7\% & 7.2\% \\
    Plain CVAE & 49.7\% & 0.5\% \\
    \bottomrule
\end{tabular}
\caption{Performance metrics for VAE-based molecule model trained for 100 epochs
  on ZINC (train) data, with and without the proposed validity model overlaid at
  test time, and the Grammar VAE method. Sample validity is the fraction of
  samples from the prior over $\mathcal{Z}$ that decode into valid molecules.}
\label{tbl:vae}
\end{table}


%% file: sections/discussion.tex
In this work we proposed a modeling technique for learning the validity
constraints of discrete spaces. The proposed likelihood makes the model easy to
train, is unaffected by the introduction of padding for variable length
sequences and, as its optimum is largely independent of the training data
distribution, it allows for the utilisation of active learning techniques.
Through experiments we found that it is vital to show the model informative
examples of validity constraints being validated. Thus, where no informative
data sets exist, we proposed a mutual-information-based active learning scheme
that uses model uncertainty to select training sequences. We used principled
approximations to make that learning scheme computationally feasible. Where data
sets of positive examples are available, we proposed a simple method of
perturbations to create informative examples of validity constraints being
broken.

The model showed promise on the Python mathematical expressions problem,
especially when combined with active learning. In the context of SMILES
molecules, the model was able to learn near-perfectly the validity of
independently perturbed molecules. When applied to the variational autoencoder
benchmark on SMILES strings, the proposed method beat the previous results by a
large margin on prior sample validity -- the relevant metric for the downstream
utility of the latent space. The model is simple to apply to existing
character-based models and is easy to train using data produced through our
augmentation method. The perturbations used do not, however, capture every way
in which a molecule may be mis-constructed. Correlated changes such as the
insertion of matching brackets into expressions are missing from our scheme.
Applying the model to a more structured
representation of molecules, for example, sequences of parse rules as used in the
Grammar VAE, and performing perturbations in that structured space is likely to
deliver even greater improvements in performance.


%% file: sections/appendix.tex

\section{Coverage estimation}
\label{appendix:coverage}
Ideally, we would like to check that the learned model $\out$ assigns positive probability to exactly those points
which may lead to valid sequences, but for large discrete spaces this is impossible to compute or even accurately estimate.
A simple check for accuracy could be to evaluate whether the model correctly identifies points as valid in a known, held-out validation or test set of real data, relative to randomly sampled sequences (which are nearly always invalid).
However, if the validation set is too ``similar'' to the training data, even showing 100\% accuracy in classifying these as valid may simply indicate having overfit to the training data: a discriminator which identifies data as similar to the training data needs to be accurate over a much smaller space than a discriminator which estimates validity over all of $\inspace$.

Instead, we propose to evaluate the trade-off between accuracy on a validation set, and an approximation to the size of the effective support of $\prod_t \out$ over $\inspace$.
Let $\inspace_+$ denote the valid subset of $\inspace$.
Suppose we estimate the valid fraction $f_+ = \nicefrac{|\inspace_+|}{|\inspace|}$ by simple Monte Carlo, sampling uniformly from $\inspace$. We can then estimate $N_+ = |\inspace_+|$ by $f_+|\inspace|$, where $|\inspace| = C^T$, a known quantity. A uniform distribution over $N_+$ sequences would have an entropy of $\log N_+$. We denote the entropy of output from the model $H$. If our model was perfectly uniform over the sequences it can generate, it would then be capable of generating $N_{\textrm{model}} = e^H$. As our model at its optimum is extremely not uniform over sequences $x \in \inspace$, this is very much a lower bound a coverage.

\begin{figure}[t]
\centering
\includegraphics[width=1.0\textwidth,trim={14cm 0 6cm 0},clip]{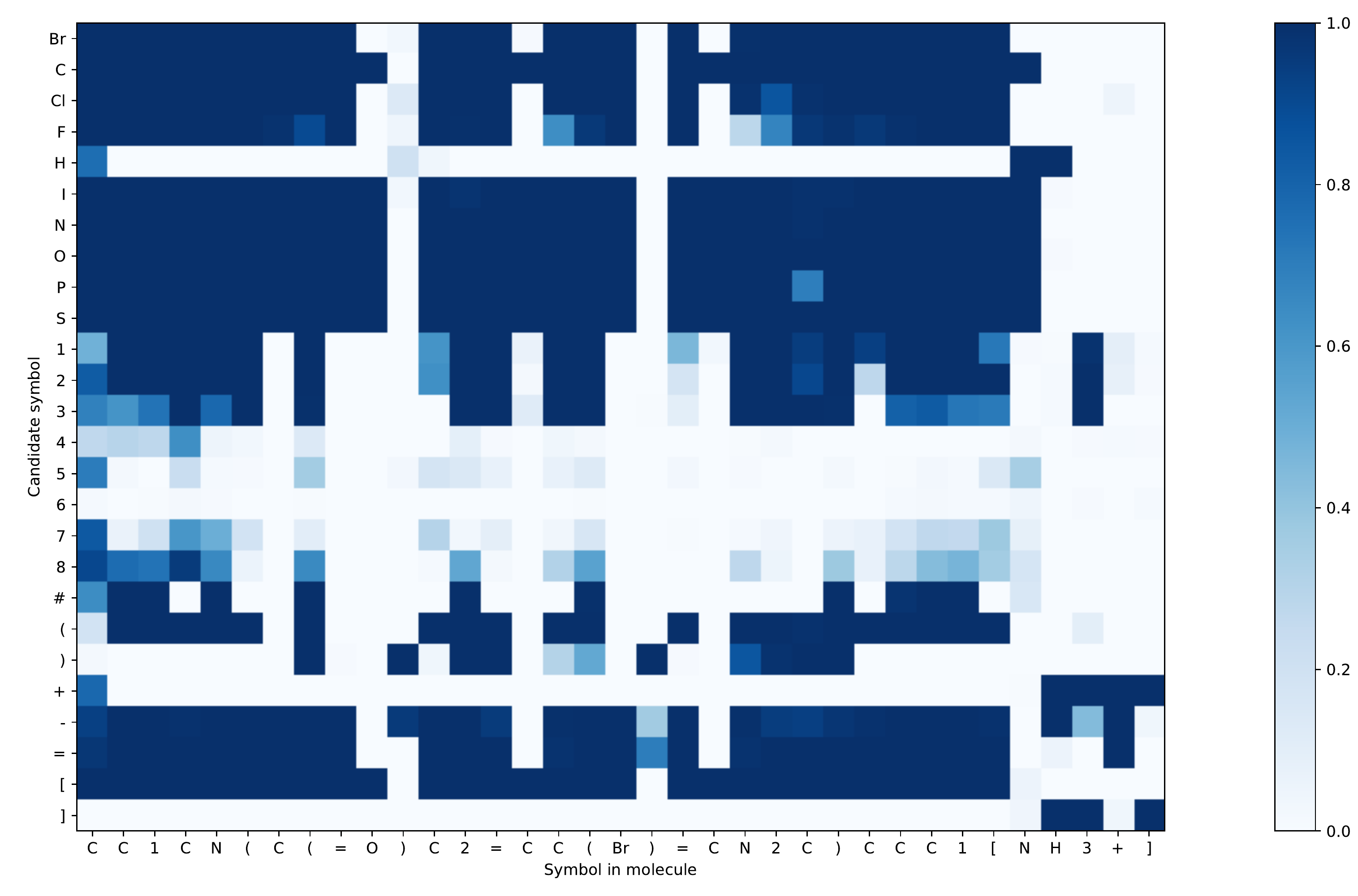}
\caption{Full heatmap showing predictions $\out$ for the molecule in Figure~\ref{fig:example-mol}.}
\label{fig:full-heatmap}
\end{figure}
